\documentclass[lettersize,journal]{IEEEtran}
\usepackage{amsmath,amsfonts}
\usepackage{algorithmic}
\usepackage{algorithm}
\usepackage{array}
\usepackage[caption=false,font=normalsize,labelfont=sf,textfont=sf]{subfig}
\usepackage{textcomp}
\usepackage{stfloats}
\usepackage{url}
\usepackage{verbatim}
\usepackage{graphicx}
\usepackage{cite}
\usepackage{booktabs}
\usepackage{subcaption}
\usepackage{multirow}
\usepackage{threeparttable}
\usepackage{makecell}
\begin{document}

\title{D4C: Improving Negative Example Quality to Enhance Machine Abstract Reasoning Ability}

\author{Ruizhuo Song, Member, IEEE,  Beiming Yuan
\thanks{This work was supported by the National Natural Science Foundation of China under Grants 62273036. Corresponding author: Ruizhuo Song, ruizhuosong@ustb.edu.cn}
\thanks{Ruizhuo Song and Beiming Yuan are with the Beijing Engineering Research Center of Industrial Spectrum Imaging, School of Automation and Electrical Engineering, University of Science and Technology Beijing, Beijing 100083, China (Ruizhuo Song email: ruizhuosong@ustb.edu.cn and Beiming Yuan email: d202310354@xs.ustb.edu.cn). }

\thanks{Ruizhuo Song and Beiming Yuan contributed equally to this work.}
}

\markboth{Journal of \LaTeX\ Class Files,~Vol.~14, No.~8, August~2021}%
{Shell \MakeLowercase{\textit{et al.}}: A Sample Article Using IEEEtran.cls for IEEE Journals}

\IEEEpubid{0000--0000/00\$00.00~\copyright~2021 IEEE}

\maketitle

\begin{abstract}

This paper is dedicated to addressing the challenge of enhancing the abstract reasoning capabilities of AI, particularly for tasks involving complex human concepts. We introduce Lico-Net, a novel reasoning engine grounded in deep learning theory, which encodes the logical structure of Raven's Progressive Matrices (RPM) problems into probabilistic representations. Lico-Net excels in solving RPM tasks. Furthermore, we propose Lico-Net-Bongard, a tailored version of Lico-Net for the Bongard-Logo problem, which also achieves high reasoning accuracy through probabilistic representations. However, we observe a mismatch between the way deep learning algorithms and humans induce reasoning concepts, primarily attributed to the inadequate quality of negative samples. Improper configuration of negative samples can convey erroneous conceptual information to deep learning algorithms, thereby distorting their learning objectives. To address this issue, we propose two novel approaches: first, treating different sample points within reasoning problems as mutual negative samples to alter the existing negative sample structure in the data; second, designing a negative sample generator based on a step-wise linear attention mechanism to produce high-quality negative samples. Experimental results demonstrate that these methods significantly improve the performance of Lico-Net (-Bongard) and other baseline models on the RPM and Bongard-Logo datasets, as well as in the domain of foundational vision model processing, particularly when addressing the NICO dataset's distribution shift problem. Our findings emphasize the importance of improving negative sample quality for enhancing the abstract reasoning capabilities of deep learning algorithms and suggest that systems represent a promising direction for future research in this field.
\end{abstract}


\section{Introduction}

\IEEEPARstart {D}{eep} learning, mimicking the human brain, has excelled in numerous fields, particularly in generative tasks like image, text, and video creation through technologies\cite{learning system} such as Generative Adversarial Networks (GANs) \cite{GAN}, Variational Autoencoders (VAEs) \cite{VAE}, autoregressive models, and Transformer \cite{Transformer}. It has also achieved significant success in computer vision \cite{ImageNet, ResNet}, natural language processing \cite{Transformer, GPT-3}, generative model construction \cite{GAN, VAE, DiffusionModel}, visual question answering systems \cite{VQA, CLEVERdataset}, and visual abstract reasoning \cite{RPM, Bongard1, Bongard2}. However, challenges remain in discriminative tasks, where deep learning struggles to fully understand human concepts embedded in the data despite some achievements\cite{context}.

This dilemma arises from deep learning models' over-reliance on superficial statistical patterns, neglecting the deeper structures and logical relationships within human concepts. Despite the tremendous success of deep learning, a significant gap persists in understanding and mimicking complex, ambiguous human concepts, especially in data-rich scenarios\cite{Human-in-the-loop}. Our research will focus on discriminative issues involving deep and complex human concepts, aiming to explore the root causes of models' poor performance and develop effective solution, thereby facilitating new advancements in deep learning's understanding and simulation of human concepts.

\section{Problem description}

Traditional visual discriminative tasks assess the generalization and feature recognition capabilities of deep learning algorithms by evaluating their ability to summarize pixel arrangements in image data. However, in this section, we introduce several non-traditional advanced visual discriminative tasks, which effectively measure the algorithms' ability to learn complex and abstract patterns underlying visible features.

\subsection{Raven's Progressive Matrices}

The Raven's Progressive Matrices (RPM) problem\cite{RPM} serves as a prestigious benchmark for assessing advanced discriminative capabilities of deep learning algorithms, as it necessitates abstract reasoning, pattern recognition, and problem-solving skills. As a high-level non-verbal task, RPM requires algorithms to possess the ability to generalize and transfer knowledge, thereby challenging them to exhibit cognitive functions akin to human intelligence. Its standardized nature and high correlation with human intelligence measurements render RPM a relevant and consistent test for evaluating the complexity of machine learning models, underscoring its authority in gauging the intelligence level of deep learning algorithms.

\subsubsection{RAVEN and PGM}

\begin{figure}[htp]\centering
	\includegraphics[trim=11cm 0cm 0cm 0cm, clip, width=8.5
 cm]{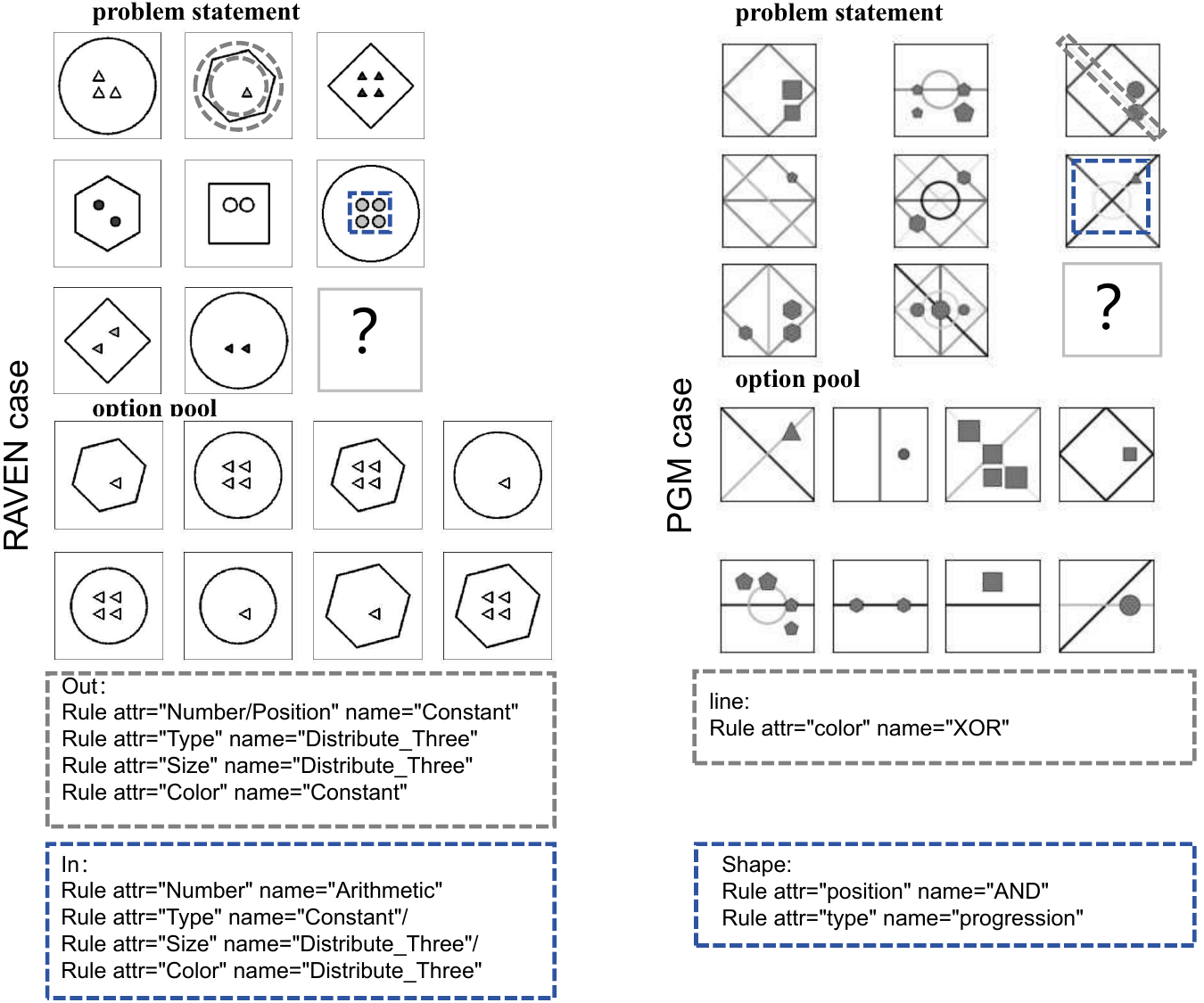}
	\caption{RAVEN and PGM case}
\label{RAVEN}
\end{figure}

The RAVEN database is a collection of Ravens Progressive Matrices problems, each consisting of 16 images. Eight of these images, known as the ``statement," are the main challenge, while the other eight form a pool for option. The goal is to choose the right images to complete a $3\times3$ matrix with a progressive geometric pattern, following an abstract ``rule." The left side of Figure \ref{RAVEN} shows an example of RAVEN. Similarly, PGM provides eight images for the problem statement and options, but its ``rule" applies to both rows and columns. The right side of Figure \ref{RAVEN} gives a typical example of PGM.

\subsection{Bongard problem}

The Bongard problem\cite{Bongard1, Bongard2} is a challenging abstract visual reasoning task that evaluates pattern recognition using minimal training examples, presenting images that follow one of two contrasting rules. It requires abstraction and learning from limited data, posing a significant challenge for both humans and machine learning algorithms. Despite its recognition in AI and cognitive science, there is limited literature on automatically solving these problems, which involve higher cognitive functions like abstract reasoning and concept learning, where machines significantly lag behind humans.

\subsubsection{Bongrad-logo}

The Bongard-Logo\cite{Bongard2} problem extends the Bongard problem, serving as a benchmark for human-level concept learning and reasoning. It captures three key aspects of human cognition: context-dependent, analogical, and sample-efficient perception with vast vocabulary. Using the LOGO language, it generates thousands of instances programmatically, challenging standard image classification methods. Experimental results show that deep learning methods fall short, highlighting the need for a more human-like perception. Bongard-Logo problems guide research towards a general visual reasoning architecture that aims to surpass this benchmark.

\begin{figure}[htp]\centering
	\includegraphics[trim=8cm 0cm 5cm 0cm, clip, width=8.5
 cm]{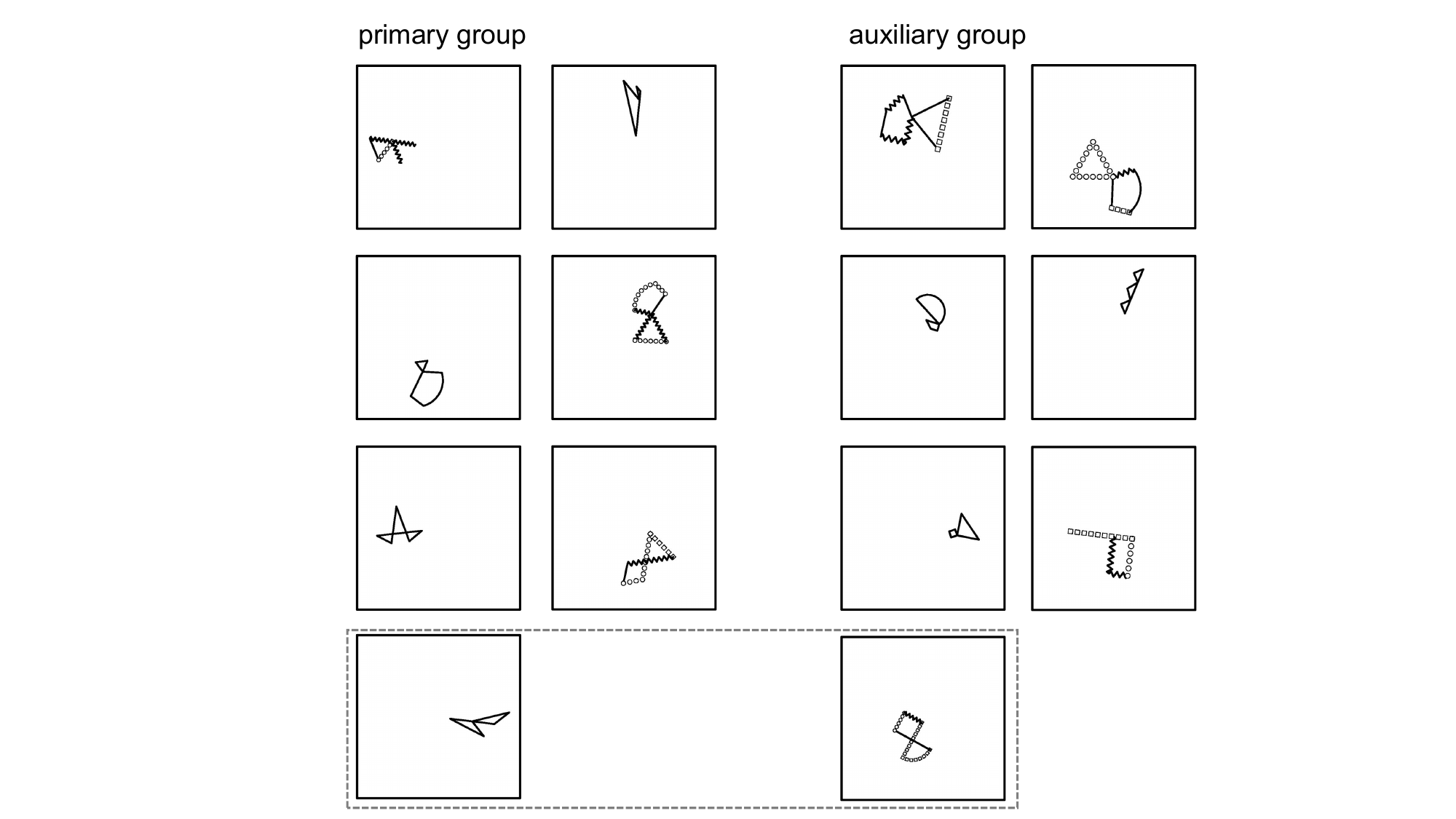}
	\caption{Bongard-Logo case}
\label{Bongard}
\end{figure}

Figure \ref{Bongard} illustrates a representative Bongard-logo case, where every Bongard case consists of two sets of images: Set A (primary) and Set B (auxiliary). Set A contains six images, with geometric entities in each image adhering to a particular set of rules, while Set B encompasses six images that do not conform to the rules observed in Set A. The task entails assessing whether the images in the test set align with the rules manifested in Set A. The level of difficulty varies depending on the case's structural complexity.
Bongard-logo problems can be categorized into three parts based on concept domains: 1) Free-form problems (FF), featuring shapes made of random action strokes; 2) Basic shape problems (BA), recognizing pattern of shape combinations; 3) High-level concept problems (HD), assessing abstract concept reasoning, like convexity and symmetry; 4) Combinatorial Abstract Shape (CM)  evaluates logical reasoning by testing systems' ability to recognize new attribute combinations not encountered in training; 5) The Novel Abstract Shape (NV) uses attribute withholding to assess innovative potential, testing models' capacity to infer new rules when an attribute and its combinations are excluded from training.

Both RPM problems and Bongard-Logo problems are challenging for humans with extensive prior knowledge, and deep learning algorithms face even higher cognitive demands in the absence of prior knowledge. Deep learning algorithms that stand out in such environments typically bring methodological breakthroughs.

\subsection{Out-of-distribution problems in image backgrounds}
Out-of-distribution (O.O.D.) problems in image backgrounds involve the challenge of identifying images with backgrounds that differ significantly from the training data, which is crucial for maintaining the reliability of machine learning models in real-world applications. These problems are particularly relevant in high-stakes fields such as medical diagnosis and autonomous driving, where misclassification can have severe consequences. Addressing O.O.D. issues often relies on learning representations that distinguish in-distribution data from O.O.D. samples, with techniques like Masked Image Modeling showing promise in improving detection performance by capturing the intrinsic distributions of the training data.

\subsubsection{NICO database}
The NICO database\cite{NICOdataset} is specifically designed to address the issues of Non-I.I.D. (Non-Independent and Identically Distributed) or Out-of-Distribution (O.O.D.) image classification problems. It simulates scenarios where the test distribution may deviate arbitrarily from the training distribution, violating the traditional I.I.D. assumption of most machine learning methods. The core idea of the NICO dataset is to label images with a main concept (e.g., dog) and context (e.g., grass), and by adjusting the proportion of different contexts in training and test data, it flexibly controls the degree of distribution shift and enables research under various Non-I.I.D. settings.

\begin{figure}[ht]\centering
	\includegraphics[width=8.5cm]{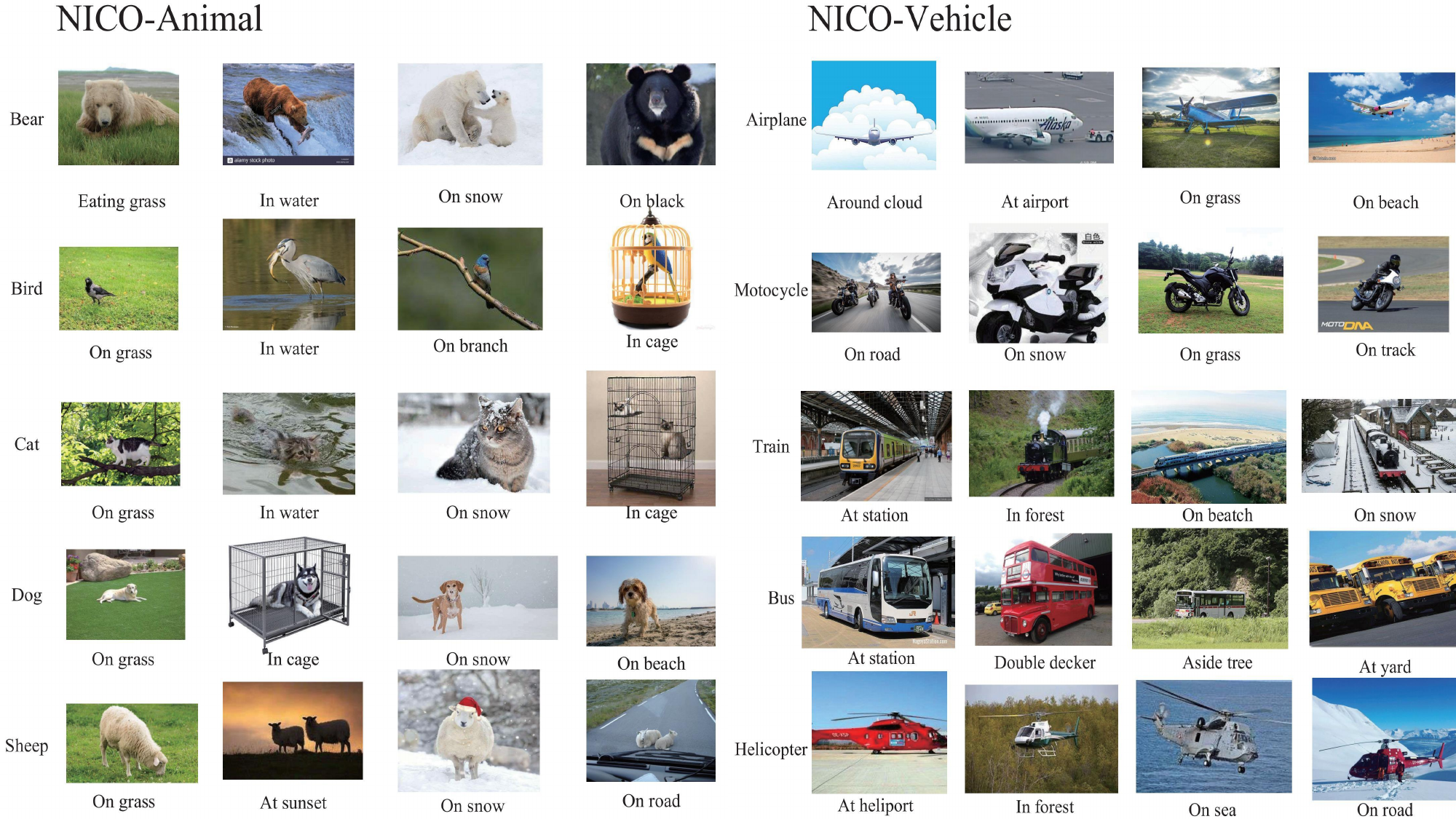}
	\caption{The figure shows the example of NICO-animal and NICO-vehicle.}
\label{vehicle}
\end{figure}

The NICO dataset includes two main categories: NICO-animals and NICO-vehicles, with 10 animal categories and 9 vehicle categories, each having 9 or 10 contexts.  Some examples of the NICO-animal and NICO-vehicle are shown in Figure \ref{vehicle}.
This paper is dedicated to exploring a deep learning approach that does not rely on pre-training or contextual prompts.



\section{Related work}

\subsection{RPM solver}

Image reasoning models such as CoPINet\cite{CoPINet}, LEN+teacher\cite{LEN}, and DCNet\cite{DCNet} focus on learning disparities and rules. Meanwhile, NCD\cite{NCD}, SCL\cite{SCL}, SAVIR-T\cite{SAVIR-T}, and neural symbolism systems (PrAE, NVSA, ALANS\cite{PrAE,ALANS,NVSA}) enhance both interpretability and accuracy. RS-CNN and RS-TRAN\cite{RS} excel in solving RPM problems, while Triple-CFN\cite{Triple-CFN} stands out by implicitly extracting and indexing concept and reasoning unit info, boosting reasoning accuracy. CRAB\cite{CRAB}, based on Bayesian modeling, has made strides with a tailored conservatory for the RAVEN database, though core challenges remain untackled, leaving the scientific community eager for the broader impact of these innovative methods.

\subsection{Bongard solver}
Addressing Bongard problems, approaches include language feature models, CNNs, and synthetic data. Depweg et al.\cite{Bongard1} used Bayesian reasoning with formal language but faced scalability issues. Kharagorgiev and Yun\cite{Bongard3} focused on feature extraction, and Nie et al.\cite{Bongard2} experimented with CNNs and meta-learning with limited success. PMoC\cite{PMoC} offers a probabilistic approach to assess group sample likelihoods, proving effective. Triple-CFN\cite{Triple-CFN} stands out for its consistent effectiveness on both Bongard-Logo and RPM problems. 

\subsection{NICO solver}
Stable learning \cite{ref13} addresses O.o.D. issues by reweighting samples to correct domain shifts and reduce feature correlations, boosting model performance on unseen tests. DecAug \cite{ref11} tackles the 2D O.o.D. problem in NICO data by adding domain labels and splitting the model into domain and object learning branches with orthogonal parameter updates for information decoupling, albeit at the cost of increased workload and categorization difficulty. SimCLR \cite{InfoNCE} enhances representation learning with data augmentation, a learnable nonlinear projector, and the InfoNCE loss. VICReg \cite{ViCReg} introduces three regularization terms on embedding variance for self-supervised image representation learning.

\subsection{Distribution distance measurement method}

\subsubsection{Wasserstein distance} The Wasserstein distance \cite{Wasssertein}, a metric for distribution dissimilarity, is given by:
\[
W(P,Q) = \mathbf{inf} \left( \sum |X-Y| \right)
\]
where \(P\) and \(Q\) are probability distributions, and \(X\) and \(Y\) are random variables from \(P\) and \(Q\), respectively. The distance \(|X-Y|\) can be Euclidean or another metric. It quantifies the minimum cost of transforming one distribution into another and is widely used in generative models to assess the difference between generated and real images.

\subsubsection{Sinkhorn distance} Sinkhorn distance\cite{Sinkhorn} is a metric based on optimal transport, which approximates the Wasserstein distance through the introduction of entropy regularization, leading to increased computational efficiency.

\section{Methodology}

In this study, we highlight the importance of deep learning algorithms learning probabilistic representations of complex human concepts. While probabilistic representations enhance performance, we observe a discrepancy between how algorithms and humans summarize concepts and logical patterns. The core issue lies in the inadequate configuration of negative samples, which conveys erroneous information and distorts learning objectives. Therefore, this study posits that, in addition to striving to develop deep learning algorithms capable of effectively learning from low-quality negative samples, improving the quality of negative samples is also a crucial strategy.

\section{Discriminator of Concepts (DC)}

This section introduces Lico-Net, a novel reasoning engine built upon Deep Neural Networks (DNNs), designed to tackle the Raven's Progressive Matrices (RPM) problem. By distilling complex human reasoning concepts into probabilistic representations, Lico-Net demonstrates exceptional reasoning performance. We further propose Lico-Net-Bongard, a model specifically designed for solving the Bongard Problems, which achieves outstanding reasoning accuracy through explicit probabilistic representation learning. Preliminary experiments on these models have uncovered a concept mismatch phenomenon, highlighting the limitations of deep models in learning implicit concepts from data. Additionally, using the NICO dataset, we demonstrate how variations in context distribution can undermine the models' ability to learn these concepts.

\subsection{A novel baseline, Lico-Net, for RPM problems}

The RPM problem requires participants to select an answer from an option pool to complete a $3\times 3$ progressive matrix of images. If the completed matrix follows a specific progressive pattern inferable from the problem statement, the answer is considered correct \cite{RPMInductivebias}. Therefore, a successful RPM discriminator should be able to recognize the consistency of various progressive patterns within a complete $3\times 3$ matrix.
To address this challenge and enhance the ability to discern underlying patterns, we have devised a novel approach. We propose Lico-Net (Link-Conception Network), whose core function is to assess the consistency of progressive patterns within a $3\times 3$ RPM matrix and quantify this consistency with a score. By sequentially applying each option provided in an RPM case to the matrix, we can utilize Lico-Net to evaluate the resulting progressive consistency and assign scores to each option accordingly. This method enables us to efficiently reason through and solve RPM problems. The following is an overview of the detailed design of Lico-Net.

The macro architecture of the Lico-Net we designed is presented in Figure \ref{Lico-net-frame-work}. As can be seen from the figure, Lico-Net is divided into four core modules: the Perception Module, the Reasoning Module, the Concept Extraction Module, and the Concept Consistency Evaluation Module. Figure \ref{Lico-net-frame-work} intuitively illustrates our concept; we plan to use the Perception Module to convert all images in the RPM matrix into corresponding representations. Subsequently, the Reasoning Module and the Concept Extraction Module will explore the concepts implicit in the representation matrices with missing entries and depict these concepts using probabilistic representations. Finally, through the Concept Consistency Measurement Module, we will accurately assess the consistency of these concepts. The specific design of the four modules is as follows:

\begin{figure}[ht]\centering
	\includegraphics[trim=0cm 5cm 0cm 4cm, clip, width=8.5cm]{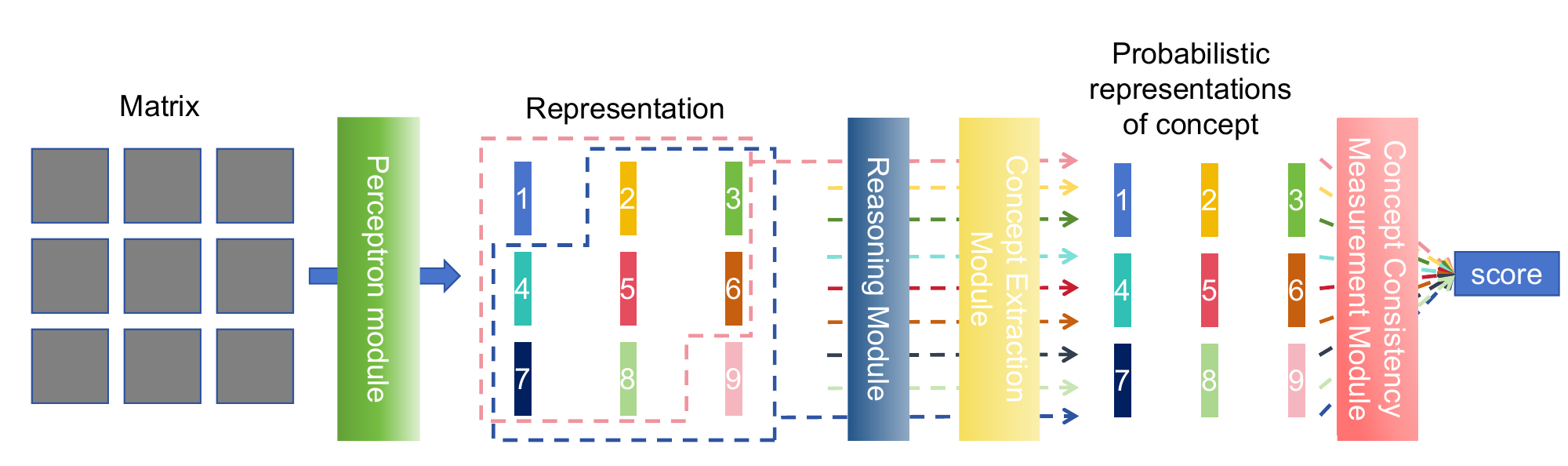}
	\caption{The figure shows the framework of Lico-Net.}
\label{Lico-net-frame-work}
\end{figure}

\subsubsection{Perceptron Module of Lico-Net}
\begin{figure}[htp]\centering
	\includegraphics[trim=0cm 4.5cm 0cm 3.5cm, clip, width=8.5
 cm]{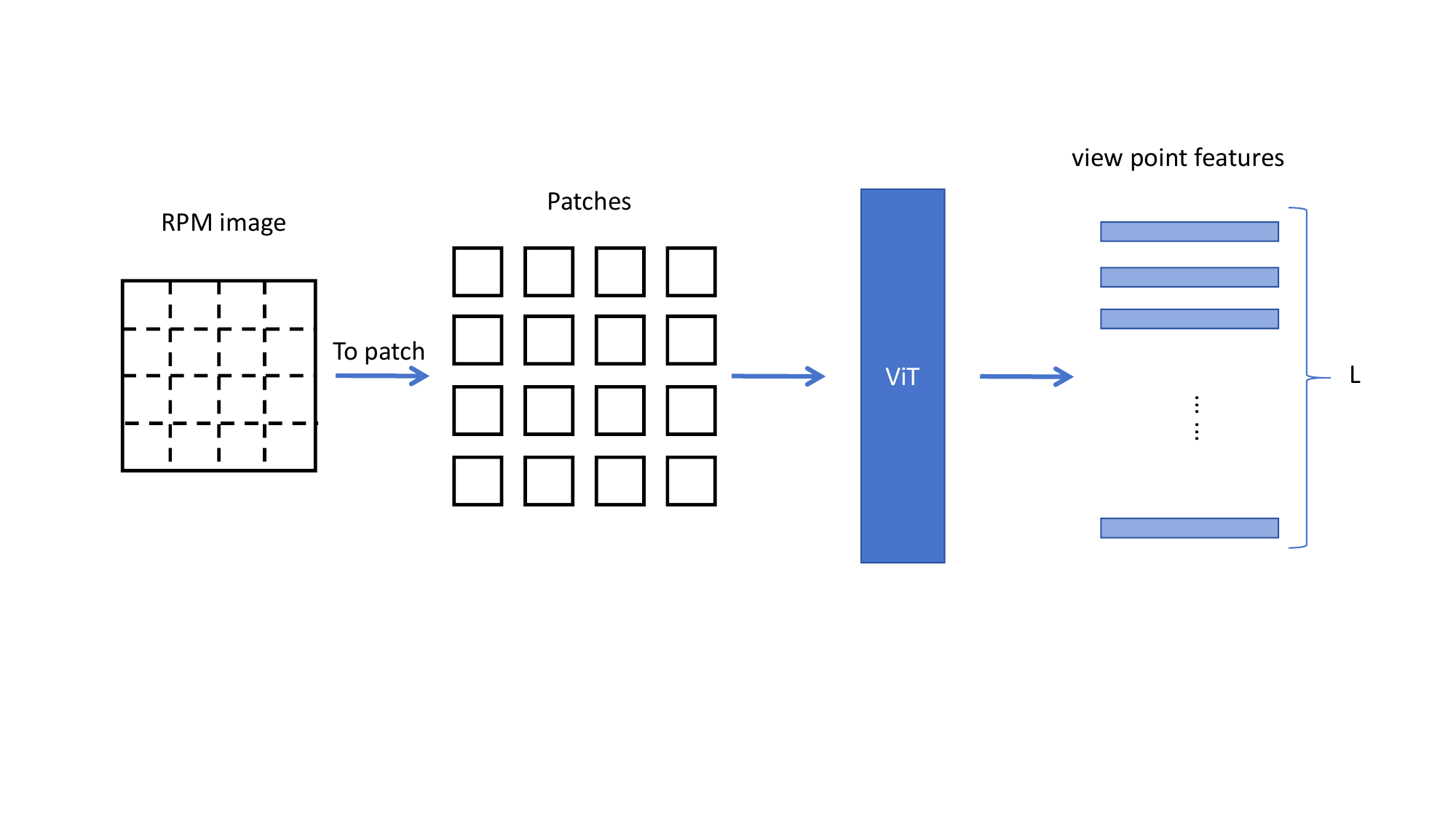}
	\caption{Perceptron of Lico-Net}
\label{Perceptron of Lico-Net}
\end{figure}

Aligned with the consensus among RPM solvers on the importance of multi-viewpoint feature extraction \cite{RS,SCL,SAVIR-T,Triple-CFN, mini pose}, Lico-Net employs the Vision Transformer (ViT) for feature extraction, retaining the entire output vectors, which are attention results, as multi-viewpoint features. We denote the number of viewpoints as $L$. Lico-Net then equally processes each viewpoint of the extracted feature. The perceptron of Lico-Net is illustrated in Figure \ref{Perceptron of Lico-Net}.

\subsubsection{Reasoning Module and Concept Extraction Module of Lico-Net}
\begin{figure}[htp]\centering
	\includegraphics[trim=0cm 0cm 0cm 0cm, clip, width=8.5
 cm]{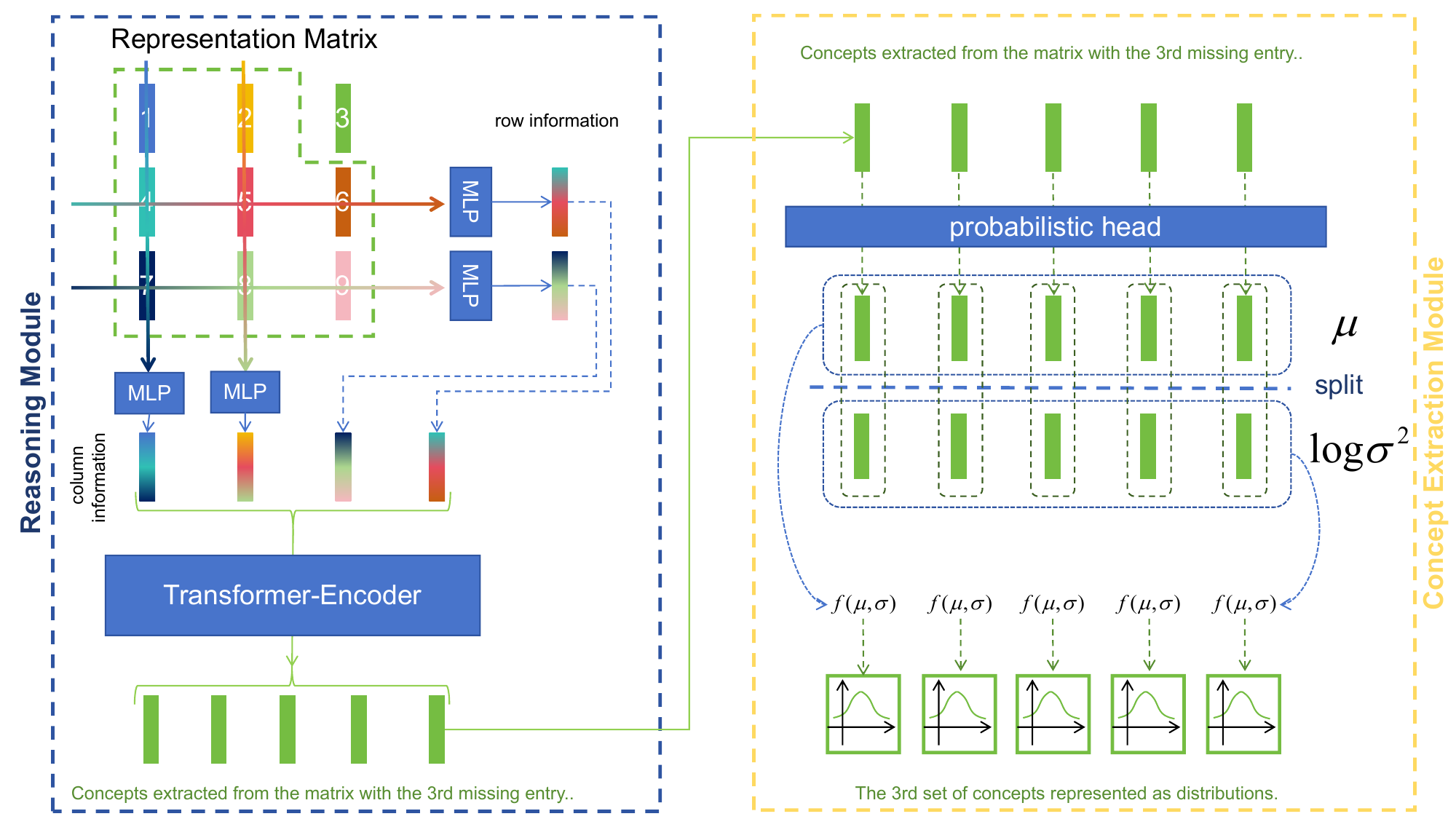}
	\caption{
The process by which the Reasoning Module handles the third incomplete representation matrix to extract concepts.}
\label{The extraction process of unit vector}
\end{figure}

The Reasoning Module and the Concept Extraction Module of Lico-Net process the representation matrices generated by the Perception Module for each viewpoint equally and in parallel, resulting in $L$ sets of concepts presented in probabilistic representations. Specifically, the Reasoning Module enumerates all possible ways in which the representation matrix can have missing entries, then extracts the concepts implied by each incomplete matrix. The Concept Extraction Module subsequently represents these concepts as probabilistic representations. We have illustrated this process in Figure \ref{The extraction process of unit vector}.

\begin{figure}[htp]\centering
	\includegraphics[trim=0cm 1cm 0cm 0.5cm, clip, width=6.5cm]{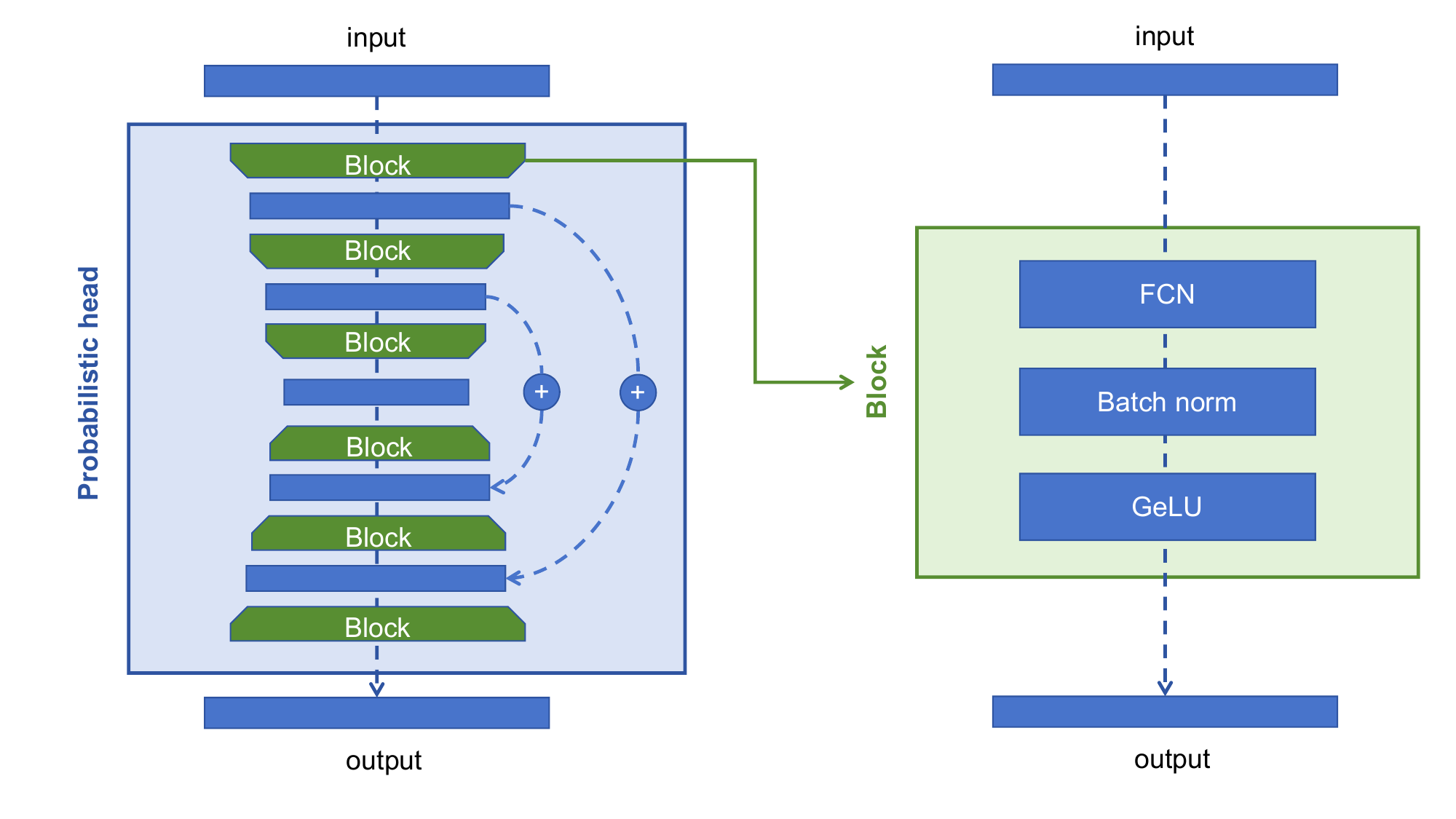}
	\caption{
The figure shows that the structure of probabilistic head.}
\label{fuzzification head}
\end{figure}

\begin{enumerate}
    \item On the left side of the figure \ref{The extraction process of unit vector}, it is easy to see that in the Reasoning Module, the MLP with a bottleneck structure extracts the remaining row and column information from the incomplete matrix, which is then fed into a standard Transformer-Encoder to extract the concepts present in the incomplete matrix. The figure shows that we use the Transformer-Encoder to extract five feature vectors from four row and column information to represent the concepts. The Transformer-Encoder capable of achieving asymmetric transformation in token counts is realized by incorporating learnable class vectors, and this technique has been successfully applied in the Vision Transformer (ViT) \cite{ViT}.
    \item  On the right side of the figure, the Concept Extraction Module is depicted, illustrating the method of converting vector-form concepts into probabilistic representation. This involves using a `probabilistic head' to double the dimensionality of the vector-form concepts, then splitting them and treating each part as the mean (\(\mu\)) and the logarithm of the variance (\(\log\sigma^2\)) of a Gaussian distribution. This achieves the transformation of vector-form concepts into distribution-form concepts. Subsequently, the reparameterization trick\cite{VAE} is employed to obtain probabilistic representations of these distribution-form concepts. 
    In this design, both the concept's baseline (\(\mu\)) and the degree of uncertainty (\(\sigma^2\)) are learned autonomously. The design of the probabilistic head is shown in Figure \ref{fuzzification head}.
\end{enumerate}

\subsubsection{Concept Consistency Measurement Module
} The Reasoning Module and the Concept Extraction Module are capable of transforming incomplete representation matrices into concepts represented by five Gaussian distributions. Samples drawn using the reparameterization trick\cite{VAE} from these five distributions are regarded as the probabilistic representations of these concepts. A complete $3 \times 3$ matrix can generate nine distinct incomplete matrices; hence, we can extract nine sets of distribution-form concepts from a complete $3 \times 3$ matrix and obtain the probabilistic representations for each set of concepts. 
The task of the Concept Consistency Measurement Module is to calculate the Sinkhorn distances \cite{Sinkhorn} among these nine distribution-form concepts, based on their probabilistic representations, and to utilize this distance to reflect the consistency score of the concepts within a complete $3 \times 3$ matrix. 
In this paper, we extract 10 probabilistic representations from each distribution-form concept to support the calculation of the Sinkhorn distance.

\subsubsection{Loss function for Lico-Net}
By employing cross-entropy loss and constraining this consistency score according to the logic of the problem, we can end-to-end optimize the three modules of Lico-Net.

\subsubsection{Preliminary experiments in Lico-Net
}
\begin{figure}[htp]\centering
	\includegraphics[trim=0cm 0cm 0cm 0cm, clip, width=6
 cm]{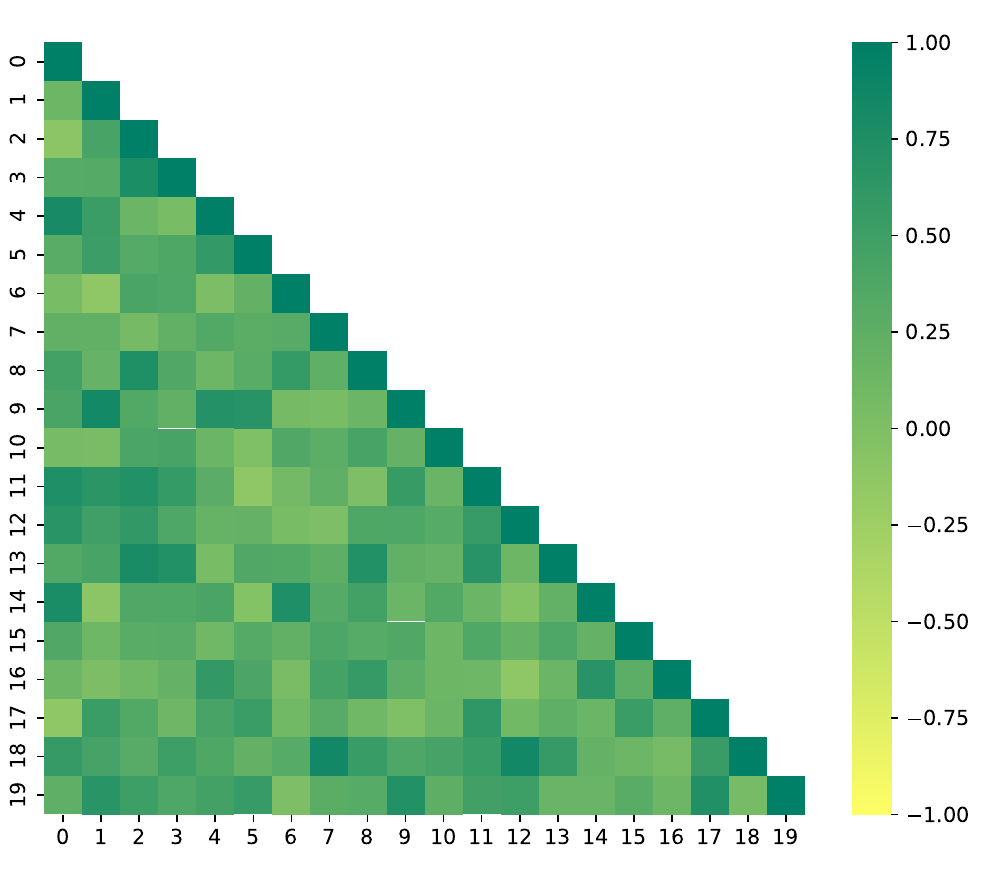}
	\caption{
The figure shows that the distances between concepts within and between these 20 problem matrices that share the same rule.}
\label{Generalization_heatmaps}
\end{figure}

After clarifying the structure of Lico-Net and its training loss function, we conducted preliminary experiments across all sub-datasets of RAVEN. Rather than focusing solely on the reasoning accuracy of Lico-Net, we placed greater emphasis on how Lico-Net learns the embedded concepts within the RAVEN dataset. Since Lico-Net can extract nine sets of concepts from a single complete $3\times 3$ matrix, it is worth delving into its concept extractor's performance across the entire dataset. Therefore, we organized this special experiment. Taking the ``Distribution 9" configuration in RAVEN as an example, 20 problem matrices with identical rules were generated for this configuration. Subsequently, we assessed the distances between concepts within and between these 20 problem matrices. As shown in Figure \ref{Generalization_heatmaps}, despite the identical rules for these 20 problem matrices, it is observed from Lico-Net's perspective that only the concepts within the same matrix are consistently matched.
In other words, Lico-Net's induction of concepts or patterns is far from human-like. Whether this pattern mismatch is a negative effect brought by probabilistic representation learning is worth discussing.

\subsection{A novel baseline, Lico-Net-Bongard, for Bongard-Logo}

The designers of Bongard-logo drew inspiration from high-dimensional human concepts, categorizing the database into three problem types: FF, BA, and HD. (The HD concept will be split into the CM and NV concepts during testing. In other words, when training the reasoning engine, the three concepts of FF, BA, and HD are involved, while during testing, the performance of the reasoning engine will be evaluated on four concepts of FF, BA, CM, and NV.) Bongard-Logo tasks participants with analyzing clustering logic between primary and auxiliary sets and categorizing images accordingly. 

We have designed a reasoning engine based on a deep neural network named Lico-Net-Bongard, which can extract abstract concepts in Bongard-Logo images and represent them as probabilistic representations. To ensure that these concepts are highly aligned with the problem-solving logic and clustering logic of Bongard-Logo, we further constrain the distances between the extracted concepts. The imposition of these constraints aims to ensure that the extracted concepts within the primary set are more closely connected, while clearly distinguishing the concepts in the auxiliary set from those in the primary set. 

\subsubsection{The structure of Lico-Net-Bongard}

\begin{figure}[htp]\centering
	\includegraphics[trim=0cm 0cm 0cm 0cm, clip, width=8.5
 cm]{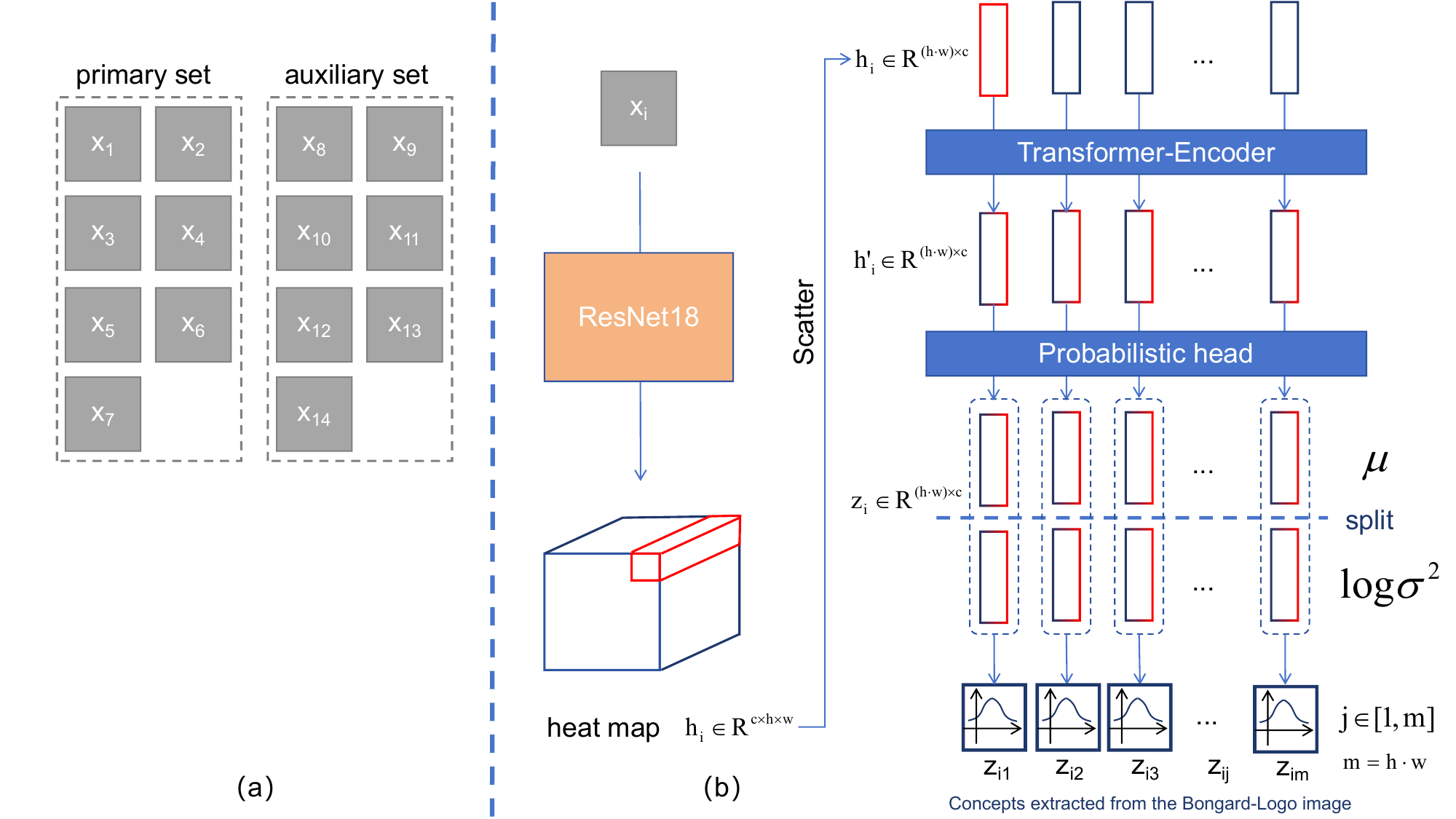}
	\caption{
The structure of Lico-Net-Bongard}
\label{The structure of Lico-Net-Bongard}
\end{figure}

For a clearer exposition of the Lico-Net-Bongard architecture, Figure \ref{The structure of Lico-Net-Bongard} has been constructed. 
\begin{enumerate}
    \item In Figure \ref{The structure of Lico-Net-Bongard} (a), we detail the annotation of the 14 images within a single Bongard-Logo problem, which is instrumental in defining the loss function for training Lico-Net-Bongard.
    \item Figure \ref{The structure of Lico-Net-Bongard} (b) illustrates the process by which Lico-Net-Bongard extracts distribution-form concepts from individual images in Bongard-Logo. The process is as follows: after utilizing ResNet18 to convert Bongard-Logo images ($x_i$) into feature maps ($h_i$), Lico-Net-Bongard scatters these feature maps along the receptive field into vector groups ($h_{ij}$ where $j$ denotes the receptive field index and $m$ represents the size of the receptive field), which are then processed by a standard Transformer-Encoder to obtain the multi-viewpoint representations ($h'_{ij}$) of the images ($x_i$). Thereafter, using a probabilistic head, these multi-viewpoint representations are individually transformed into distributions ($z_{ij}$). The samples drawn from these distributions using reparameterization techniques are regarded as the probabilistic representations of the concepts embedded in the Bongard-Logo images.
\end{enumerate}

\subsubsection{Loss function for Lico-Net-Bongard}

After Lico-Net-Bongard processes all 14 images ($\{x_i|\,i \in [1, 14]\}$) in a Bongard-Logo case to form concepts ($\{z_{ij}|\,i \in [1, 14],j\in[1,m]\}$), we can optimize Lico-Net-Bongard by constraining the Sinkhorn distances between these distribution-form concepts. The specific loss function design can be expressed as:
\begin{align}
 &{\ell _\mathbf{Lico}} \nonumber\\
    &=  \sum_{i=1}^{6}\sum_{\tilde{i}=i+1}^{7} -\log \frac{{{e^{\frac{\sum\nolimits_{j = 1}^{m}D({z_{ij}} ,{ z_{\tilde{i}j}})}{m}}}}}
    {{{e^{\frac{\sum\nolimits_{j = 1}^{m}D({z_{ij}} ,{ z_{\tilde{i}j}})}{m}}} + \sum\nolimits_{k = 8}^{14} {{e^{\frac{\sum\nolimits_{j = 1}^{m}D({z_{ij}} ,{ z_{kj}})}{m}}}} }}\nonumber\\
    &+  \sum_{i=1}^{6}\sum_{\tilde{i}=i+1}^{7} -\log \frac{{{e^{\frac{\sum\nolimits_{j = 1}^{m}D({z_{ij}} ,{ z_{\tilde{i}j}})}{m}}}}}
    {{{e^{\frac{\sum\nolimits_{j = 1}^{m}D({z_{ij}} ,{ z_{\tilde{i}j}})}{m}}} + \sum\nolimits_{k = 8}^{14} {{e^{\frac{\sum\nolimits_{j = 1}^{m}D({z_{\tilde{i}j}} ,{ z_{kj}})}{m}}}} }}\label{infoNCE}
\end{align}
The expression for \(\ell_{\mathbf{Lico}}\) is straightforward: it calculates the distances between each pair of concepts within the set \(\{z_{ij}\}_{i=1}^7\) and constrains these distances using cross-entropy loss according to the problem-solving logic. Here, \(D(\alpha, \beta)\) represents the function that calculates the Sinkhorn distance between two distributions \(\alpha\) and \(\beta\).
Similarly, calculating the Sinkhorn distance requires obtaining observed samples from the distribution, which still involves using the reparameterization trick to sample several probabilistic representations from the distribution-form concepts \(z_{ij}\), with the number of samples set to 10.

\subsubsection{Preliminary experiments in Lico-Net-Bongard
}
With the structure of Lico-Net-Bongard and its training loss function clearly defined, we conducted preliminary experiments on the Bongard-Logo dataset. We carried out two sets of experiments, one where the four types of concepts from the Bongard-Logo database were mixed together, and another where they were separated, to observe the performance of Lico-Net-Bongard.
The experimental results are shown in Table \ref{DC_Bongard_Results}.
\begin{table}[htbp]
\caption{Reasoning accuracies of Lico-Net-Bongard on Bongard-logo.}
\label{DC_Bongard_Results}
\centering
\resizebox{0.5\textwidth}{!}{
\begin{tabular}{cccccc}
\toprule
&\multicolumn{4}{c}{Test Accuracy(\%)}& \\
Model and Data Set& FF&BA&CM&NV \\
\midrule
Lico-Net-Bongard on Bongard-logo&91.0&98.3&76.8&77.0\\
\midrule
Lico-Net-Bongard on Separated Bongard-logo&98.8&99.3&75.5&73.4\\
\bottomrule
\end{tabular}
}
\end{table}
As observed in Table \ref{DC_Bongard_Results}, when the concepts in the Bongard-Logo dataset are artificially separated, Lico-Net-Bongard's performance improvement in BA and FF, which are independent and identically distributed concepts, and the performance decline in NV and CM, which are generalization concepts, both reflect the situation where Lico-Net-Bongard confuses these four concepts.

\subsection{The baseline on the NICO database.}
The NICO database\cite{NICOdataset} presents a classic visual discrimination task, but it exhibits a distribution shift between background and foreground pixels. In NICO, ``domain" refers to the background, which can undergo two types of shifts: Diversity Shift, where the test set contains background styles not seen in the training set; and Correlation Shift, where a predominant association between background and foreground in the training data leads the visual model to incorrectly associate labels with the background, resulting in decreased performance.

\subsubsection{Preliminary experiments in NICO}
From an experimental perspective based on the NICO dataset, we describe the issues of Diversity Shift and Correlation Shift.

Diversity Shift: We randomly select 8 domains as the source domains, with the remaining 2 domains serving as the target domains. We first train and test some basic deep learning models on all 10 domains, using the test accuracies as a benchmark. Then, we train the model with data from the source domains and test it on the target domains, comparing the test results with the benchmark to observe the impact of Diversity Shift. To ensure a fair comparison, we maintain consistent data quantities in the training and testing sets across all experimental setups. The experimental results shown in Table \ref{Impact_vehicle} indicate the significant impact of Diversity Shift on deep learning models.
\begin{table}[ht]
\centering
\caption{Experiment of Diversity Shift Impact (8:2)}
\label{Impact_vehicle}
\begin{threeparttable}
\begin{tabular}{cccc}
\hline\hline
dataset &model         &\makecell[c]{Domain \\mixing(10:10)\tnote{1}} &\makecell[c]{Domain \\shift(8:2)\tnote{2}}\\
\hline

\multirow{4}*{NICO-Animal}&ResNet18  	   &41.03$\pm$0.10\% &28.56$\pm$0.30\%  \\

~&ResNet50	   &54.55$\pm$0.15\%  &39.55$\pm$0.25\%  \\

~&Vgg16       	&62.49$\pm$0.21\% &48.66$\pm$0.50\% \\ 

~&Vgg19       	&60.22$\pm$0.13\% &46.11$\pm$0.09\% \\ 

\hline

\multirow{4}*{NICO-Vehicle}&ResNet18  	   &53.45$\pm$0.12\% &42.29$\pm$0.08\%  \\

~&ResNet50	   &70.20$\pm$1.18\%  &55.16$\pm$0.35\% \\

~&Vgg16       	&74.13$\pm$1.33\% &58.24$\pm$0.40\%  \\

~&Vgg19       	&74.59$\pm$1.26\% &57.60$\pm$0.80\% \\

\hline\hline

\end{tabular}

 \begin{tablenotes}
        \footnotesize
       \item[1]Separate training set and test set from mixed 10 domians.
\item[2]The training set is from the source domain, and the test set is from the target domain.
 \end{tablenotes}

\end{threeparttable}

\end{table}

Correlation Shift: In the experiment investigating the impact of Correlation Shift, we amplify this phenomenon by randomly reducing samples in some training sets. Specifically, we select one of the 8 source domains as the primary domain and reduce the number of samples in the remaining 7 domains to $1/5$ of the primary domain, making them secondary domains. We use the reduced source domains (primary and secondary) for training, while the target domain remains unchanged. The experimental results of Vgg16 on NICO with a stronger domain shift are shown in Table \ref{correlation shift impact}, which we use as a benchmark for the Correlation Shift problem.

\begin{table}[ht]
\centering
\caption{Vgg16 on nico 8:2 with Correlation shift}
\label{correlation shift impact}
\begin{threeparttable}
\begin{tabular}{cccc}
\hline\hline
\makecell[c]{dataset}&model&\makecell[c]{original \\(8:2)\tnote{1}}&\makecell[c]{Stronger domain\\ shift(8:2)\tnote{2}} \\
\hline

\makecell[c]{NICO-Animal}&\makecell[c]{vgg16}	
&48.66$\pm$0.50\%&37.83$\pm$0.12\%\\

~&~	&~&~\\

\makecell[c]{NICO-vehicle}&\makecell[c]{vgg16}
&58.24$\pm$0.40\%  	&41.85$\pm$0.25\% \\ 

\hline\hline
\end{tabular}
 \begin{tablenotes}
        \footnotesize
       \item[1]Do not change the data volume proportion of the source domain.
\item[2]The data volume of the secondary domain in the source domain is reduced by $1/5$ of the primary domain.
 \end{tablenotes}

\end{threeparttable}
\end{table}

Since prevalent pre-training (e.g., ImageNet pre-trained models) covers most of the information needed for our experiments, none of the experiments on NICO adopt a pre-training scheme. We conduct each experiment 5 times and report the average of the 5 test accuracies.

\subsection{
Discussion on Preliminary Experiments}

In our preliminary experiments, there is a pervasive phenomenon that there exists a mismatch between the way deep learning models infer deep concepts and the way humans define these concepts.

In the Bongard-Logo problem, Lico-Net-Bongard may need to store different pixel encoding patterns for different concepts. When these patterns are mutually exclusive, mixed concepts can confuse the model, leading to uncertainty in problem-solving. 
This transforms four concepts that are clear and organized from a human perspective into four conflicting concepts from the perspective of Lico-Net-Bongard.
As a result, the phenomenon of concept mismatch observed in the concept dissection experiment becomes understandable.
However, unlike the Bongard-Logo problem, the correct answers for different instances in the RAVEN problem are all designed based on the same set of inherent rules and problem-solving logic, and the annotations in NICO are also accurately aligned with the foreground subjects. Thus, the question arises: why do RAVEN and NICO also exhibit a certain degree of concept mismatch?

\section{Attribution Discussion of Mismatch in Deep-level Concepts (D2C)}

Our preliminary experiments have revealed a significant finding: there exists a mismatch between the way our designed reasoning engine induces deep or implicit concepts in the data and the way humans interpret these concepts. This mismatch is one of the key factors limiting the performance of the reasoning engine. Therefore, delving into the causes of the concept mismatch and devising corresponding solutions will be an effective approach to enhancing the performance of the reasoning engine.
Upon analysis, one of the reasons for concept mismatch is attributed to the low quality of the negative examples set in the problems.

\subsubsection{RPM problem} Taking the RPM problem, most end-to-end RPM reasoning engine, including Lico-Net, follow a pattern: when the correct option is filled into the problem matrix, the reasoning engine returns a high probability value for this matrix, and conversely, returns a low probability value.
This indicates that the essence of an RPM reasoning engine is to perform a conditional probability distribution over the options, conditioned on the problem matrix. The optimization of the reasoning engine can be seen as an adjustment of this conditional probability distribution. If we assume that the form of this conditional probability distribution, after determining the problem matrix, is a Gaussian distribution, then the optimization of this distribution can be regarded as adjusting the mean \( \mu_{\theta} \) and variance \( \sigma_{\theta}^2 \) in the following formula:

\begin{align}
    &-\log(\text{Engine}(x_\text{option}| \,\text{matrix} )) \nonumber
    \\&= \frac{{(x_\text{option} - \mu_{\theta})^{2}}}{{2\sigma^{2}_{\theta}}} - \log(|\,\sigma_{\theta}|\,) - \frac{1}{2}\log(2\pi)\label{normal_distribution}
\end{align} Here, \( x_\text{option} \) represents the options filled into the matrix, and \( \mu_{\theta} \) and \( \sigma_{\theta}^2 \) represent the mean and standard deviation of the Gaussian distribution, respectively.

It can be observed that when positive samples (correct option) are substituted into the formula \ref{normal_distribution}, we aim for the smallest possible result, which corresponds to minimizing the Euclidean distance between option $x_\text{option}$ and the mean of the correct solution $\mu_\theta$. Conversely, when negative samples (incorrect options) are substituted, we desire the largest possible result, equivalent to minimizing the standard deviation $\sigma_{\theta}^2$.
It is well understood that the mean $\mu_\theta$ describes the center of a Gaussian distribution, while variance $\sigma_{\theta}^2$ describes the spread or dispersion. Consequently, it is not difficult to comprehend that the learning objective of end-to-end RPM reasoning engine like Lico-Net is a probability distribution delimited by positive and negative examples. In other words, the learning goal of these scoring-based end-to-end reasoning engine is not the reasoning logic established by humans, but rather the distinction between positive and negative examples. 
This indicates that when the quality of negative examples is not sufficiently high or not deceptive enough, it can lead to the engine learning concepts with relatively loose boundaries, which also causes the learned concepts to deviate significantly from the reasoning logic established by humans.

\subsubsection{Bongard-Logo problem}This issue has also impacted the process of solving and learning the Bongard-Logo problem. By examining Equation (\ref{infoNCE}), it is evident that the primary learning focus of Lico-Net-Bongard remains on ensuring that the similarity among concepts extracted from positive examples is higher than that between positive and negative examples. In other words, its learning objective is to distinguish the differences between positive and negative examples, rather than directly grasping the four specific concepts defined by humans in Bongard-Logo. Consequently, due to the inadequate quality of the negative examples provided by Bongard-Logo, it communicates a set of conflicting concepts.

\subsubsection{NICO problem} In the NICO problem, due to the special data distribution, traditional visual deep neural networks also tend to overly focus on background information and elements when learning foreground object classification tasks. In the NICO dataset, each type of subject has its unique domain, and deep networks optimized using cross-entropy loss inherently concentrate on learning the distinctions between positive and negative examples. Therefore, it is understandable that the background, as a prominent distinction between positive and negative examples, is emphasized and learned by the deep networks.

\subsubsection{Summary}In summary, most end-to-end deep learning models focus more on the differences between positive and negative examples in the data, rather than the reasoning logic behind why positive examples are correct and negative examples are incorrect. When the quality of negative examples in the data is not high enough, the distinction between positive and negative examples is unable to accurately convey the underlying logic and embedded concepts.

\section{Define New Negative Examples to Delineate the Deep-level Concept (D3C)}

Previous works, including Lico-Net(-Bongard), have primarily focused on designing deep learning network architectures that can achieve high reasoning accuracy on low-quality negative examples. However, this paper introduces new perspectives and a novel approach to addressing this issue.

Given that we have discovered the quality of negative examples is crucial for deep learning networks to learn the deep concepts within data, we can set more and higher-quality negative examples for our target data to enhance the performance of deep learning models. Therefore, the methodology of this section is quite straightforward: when a deep network learns from an individual observed sample in the data, we integrate other observed samples as negative samples into the current sample.

\subsection{RPM and Bongard-Logo}
\begin{figure}[htp]\centering
	\includegraphics[trim=0cm 0cm 0cm 0cm, clip, width=8
 cm]{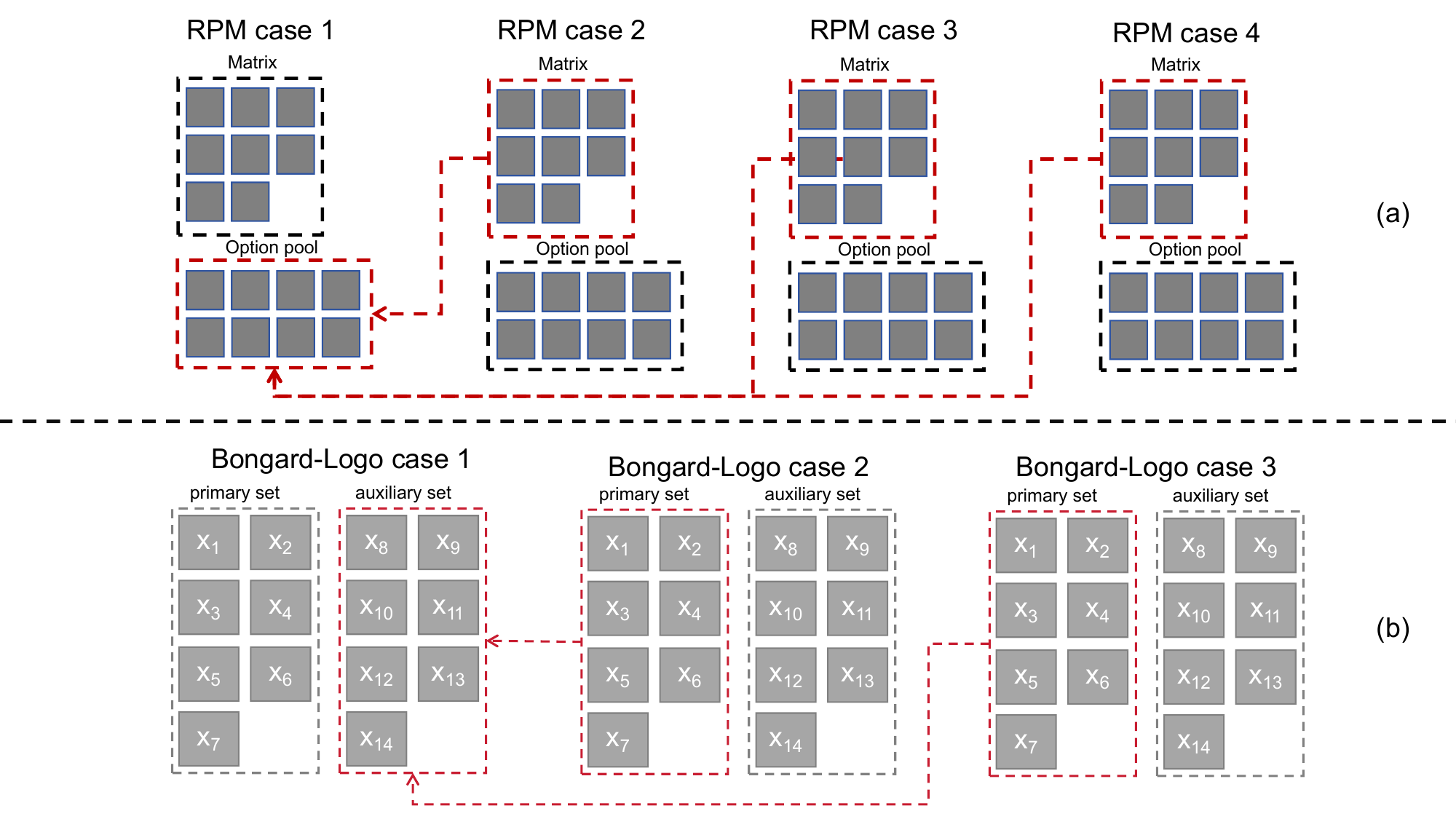}
	\caption{
The structure of Lico-Net-Bongard}
\label{cross-neg_instance}
\end{figure}

When tackling the RPM and Bongard-Logo problems, we enhance the option pool of a given observed sample by selecting images from two other distinct observed samples. The specific procedure is shown in Figure \ref{cross-neg_instance}. 
Figure 9(a) outlines the approach for RPM problems, and Figure 9(b) illustrates the method for Bongard-Logo problems. Due to computational limitations, we augment the negative examples with three extra samples for RPM and two extra samples for Bongard-Logo.

\subsection{NICO}

For the NICO dataset, we cannot use external images as negative samples because it would disrupt the data distribution and diminish its significance. This is also why we avoid importing pre-trained model parameters. However, our methodology for improving the quality of negative samples remains unchanged.

When using a basic deep learning model to process NICO data, positive and negative sample comparison is implicitly done through cross-entropy. To enhance the quality of negative samples without extra data or supervision, this paper proposes an explicit comparison method: the Group-based Step-wise Comparison Learning Method (GSCLM). The design of GSCLM is quite straightforward. It separates the training of the perceptual part (convolutional layers) and the inductive part (fully connected layers) of the visual deep learning model, such as Vgg16. 

\subsubsection{The training of the perceptual part}
\begin{figure}[ht]\centering
	\includegraphics[width=8.5cm]{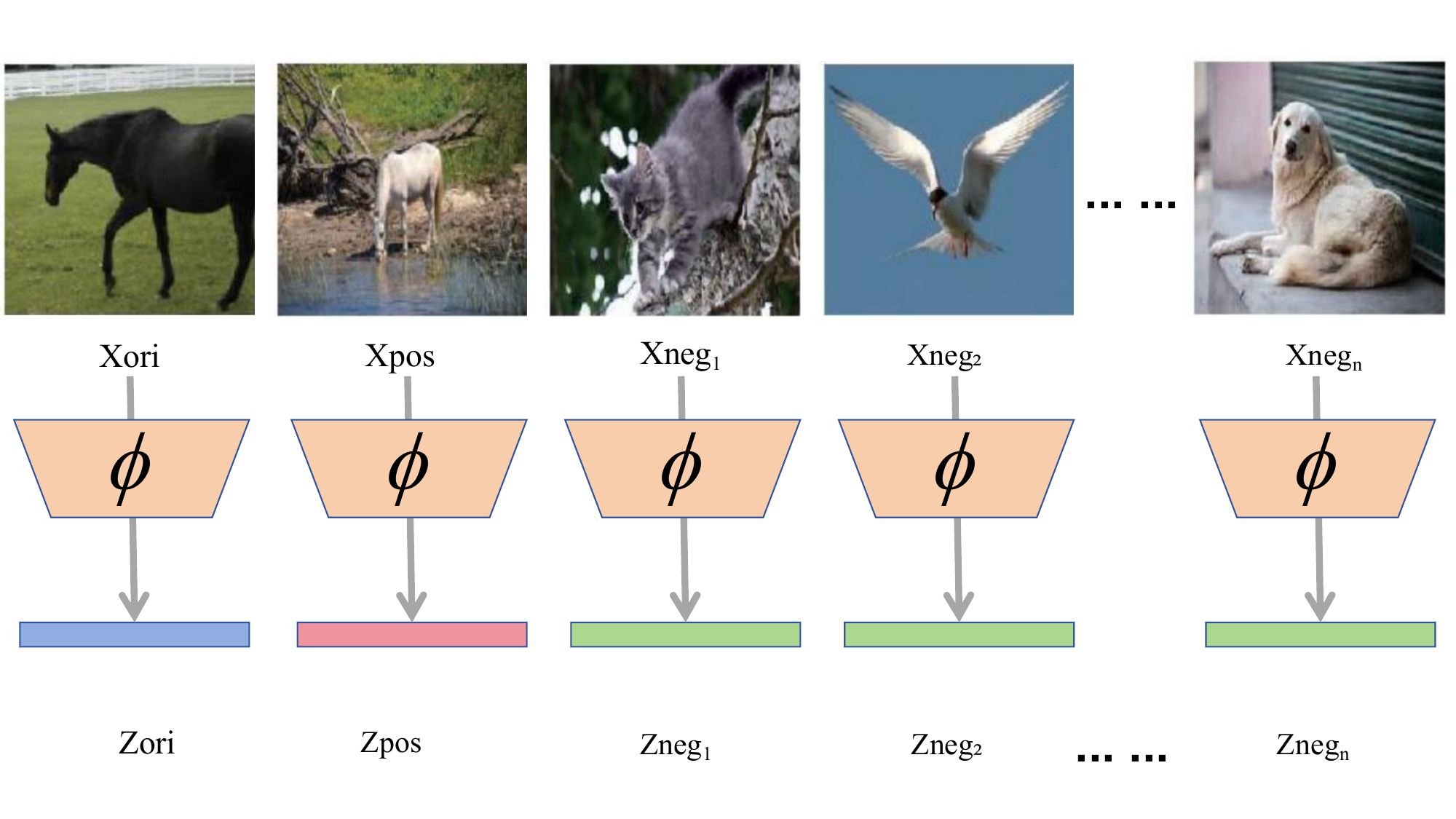}
	\caption{An instance of a representation set.}
\label{loss_function}
\end{figure}

After the perceptual part of a deep learning network processes a NICO image $x_{\text{ori}}$ into its representation $z_{\text{ori}}$, this paper selects a representation $z_{\text{pos}}$ of a same-category image $x_{\text{pos}}$ and representations $\{z_{\text{neg}_m}\}_{m=1}^M$ of multiple different-category images $\{x_{\text{neg}_m}\}_{m=1}^M$ to form a representation group. In Figure \ref{loss_function}, we provide an example of a representation group.
In the figure, ``$\phi$" represents the perceptual part of a visual deep learning network.
Subsequently, we impose loss constraints on this representation group to optimize the perceptual component of the deep network. The loss constraints can be expressed as:
\begin{align}
    &{\ell _\mathbf{D3C}}({z_{pos}},{ z_{ori}},\{ {z_{ne{g_m}}}\} _{m = 1}^M) \nonumber\\
    &=  -\log \frac{{{e^{({z_{pos}} \cdot { z_{ori}})/\tau }}}}{{{e^{({z_{pos}} \cdot { z_{ori}})/\tau}} + \sum\nolimits_{m = 1}^M {{e^{({z_{pos}} \cdot {z_{ne{g_m}}})/\tau}}} }}\nonumber\\
    &-\log \frac{{{e^{({z_{pos}} \cdot { z_{ori}})/\tau}}}}{{{e^{({z_{pos}} \cdot { z_{ori}})/\tau}} + \sum\nolimits_{m = 1}^M {{e^{({z_{ori}} \cdot {z_{ne{g_m}}})/\tau}}} }}\nonumber\\
    &+{\ell _{{\rm{cov}}}}\left(\{{z_{pos}}\cup{ z_{ori}}\cup \{ {z_{ne{g_m}}}\} _{m = 1}^M\} \right)\label{infoNCE-nico}
\end{align}
Where $\tau$ is a temperature coefficient with a value of 0.01, $M$ represents the number of negative samples in the representation group, which is set to 9 in this paper, and ${\ell _{{\rm{cov}}}}$ is a common loss constraint term for the correlation between dimensions within the representation, which is utilized in VICReg\cite{ViCReg}.

The design of these loss constraints may resemble traditional contrastive learning algorithms; however, it is crucial to highlight that, in contrast to the emphasis on a large number of negative samples in SimCLR and the large batch learning in ViCReg, the representation group contrast approach proposed in this paper requires only a minimal number of negative samples for participation.

\subsubsection{The training of the inductive part}
After the perceptual component of the deep network is well-trained, we freeze these parameters and subsequently begin training the inductive part of the deep network.

\section{Designing More Demanding Negative Examples to Drive Deep-Level Concept Learning (D4C)}
The methodology presented in the D3C section for RPM problems and Bongard-Logo problems is undoubtedly rudimentary. Compared to the methodology for the NICO problem, its effectiveness in improving the quality of negative examples remains uncertain. Whether and to what extent the new negative examples introduced by D3C for RPM and Bongard-Logo problems advance the development of deep concepts is still uncertain. Therefore, this paper proposes a more reasonable approach.

\begin{figure}[ht]\centering
	\includegraphics[width=7.5cm]{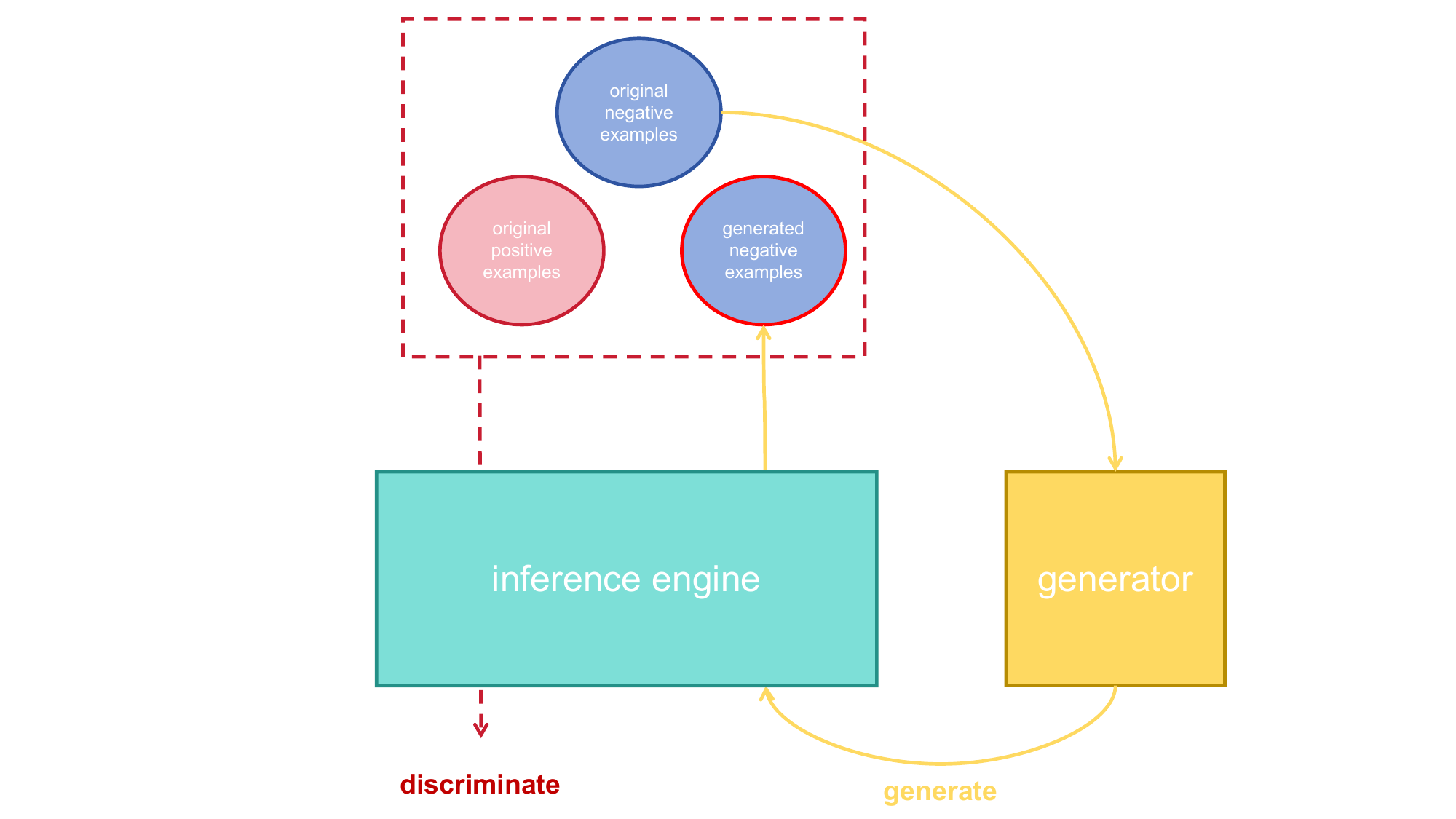}
	\caption{The framework that integrates the generator with the reasoning engine}
\label{gan}
\end{figure}

We have recognized that the low quality of negative examples hinders deep models from learning precise concepts. To obtain higher-quality negative examples without increasing data or supervision, we propose leveraging the probability feedback provided by the reasoning engine as an evaluation metric to guide the generation of better negative examples. Therefore, a sample generator that produces negative examples with higher evaluations from the reasoning engine is introduced, and its architecture is shown in Figure \ref{gan}.
As can be seen from the figure, we have adopted the traditional framework of adversarial generation but with key modifications. By defining the structure of the generator, we can present our methodology.

\begin{figure}[ht]\centering
	\includegraphics[width=8.5cm]{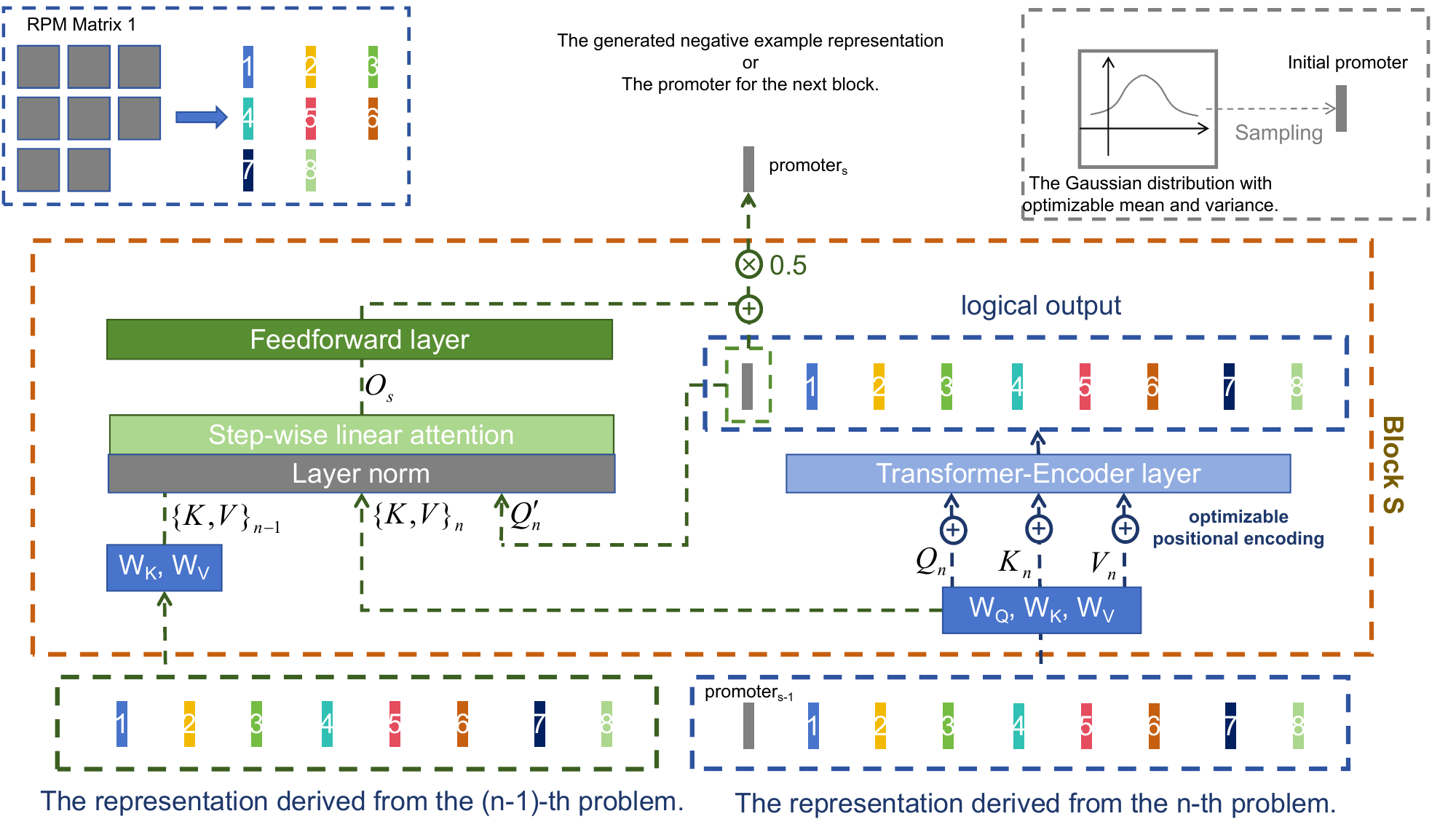}
	\caption{The figure illustrates one block of the negative example representation generator designed in this paper.}
\label{generator}
\end{figure}

It is important to note that negative examples are not unique, and consequently, our negative example generator must function as a one-to-many mapping. This poses a challenge for traditional deep learning structures to effectively perform this task, as one-to-many mappings are non-differentiable and cannot be optimized using conventional gradient descent algorithms. Therefore, this paper designs a representation generator as shown in the Figure \ref{generator}. 
Figure \ref{generator}, taking RPM problems as an example, demonstrates the basic block structure of the negative example generator designed in this paper. It is evident that this structure is also applicable to generating negative examples for Bongard-Logo and NICO problems.

As illustrated in the Figure \ref{generator}, our negative example generator does not directly generate images; instead, it generates representations of negative examples.
The figure shows that after combining the representations of all images in an RPM problem matrix with a promoter and mapping them into query, key, and value vectors ($Q_n, K_n, V_n$), we use a single-layer Transformer-Encoder to compute the logical connections between them. Subsequently, within the logical output of the Transformer-Encoder, the representation corresponding to the promoter ($Q'_n$) is utilized as the query to compute the step-wise linear attention mechanism with the key-value pairs formed by the representations of all images in other RPM problem matrices ($K_{n-1}, V_{n-1}$). The computation process of Step-wise linear attention mechanism can be represented as:
\begin{align}
{\text{O}_s} &= \frac{\sigma(Q'_n)M_{n-1}}{\sigma(Q'_n)z_{n-1}} \label{Q}\\
M_n &= M_{n-1} + \sigma(K_n)^T\left(V_n - \frac{\sigma(K_n)M_{n-1}}{\sigma(K_n)z_{n-1}}\right) \label{M}\\
z_n &= z_{n-1} + \sum \sigma(K_{n})\label{Z}
\end{align}
Formula (\ref{Q}) implicitly involves the participation of $K_{n-1}$ and $V_{n-1}$. In this paper, we use the element-wise ELU + 1 function as the activation function $\sigma(\cdot)$. After the step-wise linear attention mechanism calculation is completed, the resulting $O_s$ is processed by a feedforward layer, which follows the classic design of the Transformer. This processed output is then summed with $Q'_n$ to obtain the updated promoter. The updated promoter can either serve as the representation of the generated negative example or act as the input promoter for the block stacked after the current block.

Our architecture adopts a pattern similar to the infinite attention mechanism and incorporates an additional memory bank based on the linear attention mechanism to store the representations of images from the training set. As training iterations proceed, the step-wise linear attention mechanism continuously stores the image representations read during previous training batches. It is worth noting that during the first training batch, we do not compute \(O_1\), but we still update \(M_1\) and \(z_1\). The update process can be explained as follows:
\begin{align}
M_1 &= \sigma(K_1)^TV_1  \\
z_1 &= \sum \sigma(K_{1})
\end{align}

After stacking the generator blocks depicted in Figure \ref{generator} $S$ times without sharing parameters, a complete generator is formed. An initial prompter, ${Prompter}_0$, sampled from a normal distribution with optimizable mean and variance, is progressively updated through each of the $S$ blocks in the generator until it evolves into ${Prompter}_S$. This final ${Prompter}_S$ is utilized as the generated negative representation.
It is not difficult to see that the generator designed in this paper has an advantage: variations in the initial promptor can lead to diversity in the generated negative examples. 

During the training of the generator, it is optimized alternately with the target reasoning engine, with the expectation that the generated negative examples score higher in the engine than the original negative examples. This paper employs the cross-entropy loss function to achieve this learning objective of the generator, enabling it to provide tighter concept boundaries for the reasoning engine, thereby achieving a performance breakthrough for the already trained reasoning engine. Based on the principle that unsupervised learning does not ensure disentanglement \cite{unsupervised learning of disentangled representations}, the design of such a loss function indicates that this paper is not concerned with the specific attributes or concepts conveyed by the generated negative examples.

\section{Experiment}
All experiments conducted in this paper were programmed in Python, utilizing the PyTorch\cite{Pytorch} framework.

\subsection{Experiment on  NICO}
This paper conducted experiments on NICO, which included exploring the performance changes of ResNet18, ResNet50, VGG16, and VGG19 when combined with the D3C and D4C algorithms. 

D3C enhances the positive-negative example contrast process in traditional visual deep learning models and thereby improves the quality of negative examples. Therefore, to attach D4C to these visual models, D3C must serve as the foundation. When using the D4C method to provide high-quality negative examples for the visual model processing NICO data, we maintain a memory bank for each category in the NICO dataset, that is, we maintain an $M_{n-1}$ and $Z_{n-1}$ for each category. When generating high-quality negative example representations for a specific subject category in NICO, we input one sample representation from each of the other categories into the negative example generator. That is, when generating higher-quality negative examples for $z_{ori}$, we require the $\{z_{\text{neg}_m}\}_{m=1}^M$ selected during the establishment of the representation group in the D3C method to serve as inputs to the negative example generator. Then, the corresponding memory banks $M_{n-1}$ and $Z_{n-1}$ are selected based on the subject category to assist in the generation process. During training, we utilized the Adam\cite{ADAM} optimizer with a learning rate of $10^{-3}$. The Adam optimizer contains a common deep network optimization method.
\cite{optimization}. The experimental results are recorded in Table \ref{D34C_nico1}. D4C method generates 5 negative examples for each NICO sample. 

\begin{figure}[ht]\centering
	\includegraphics[width=8cm]{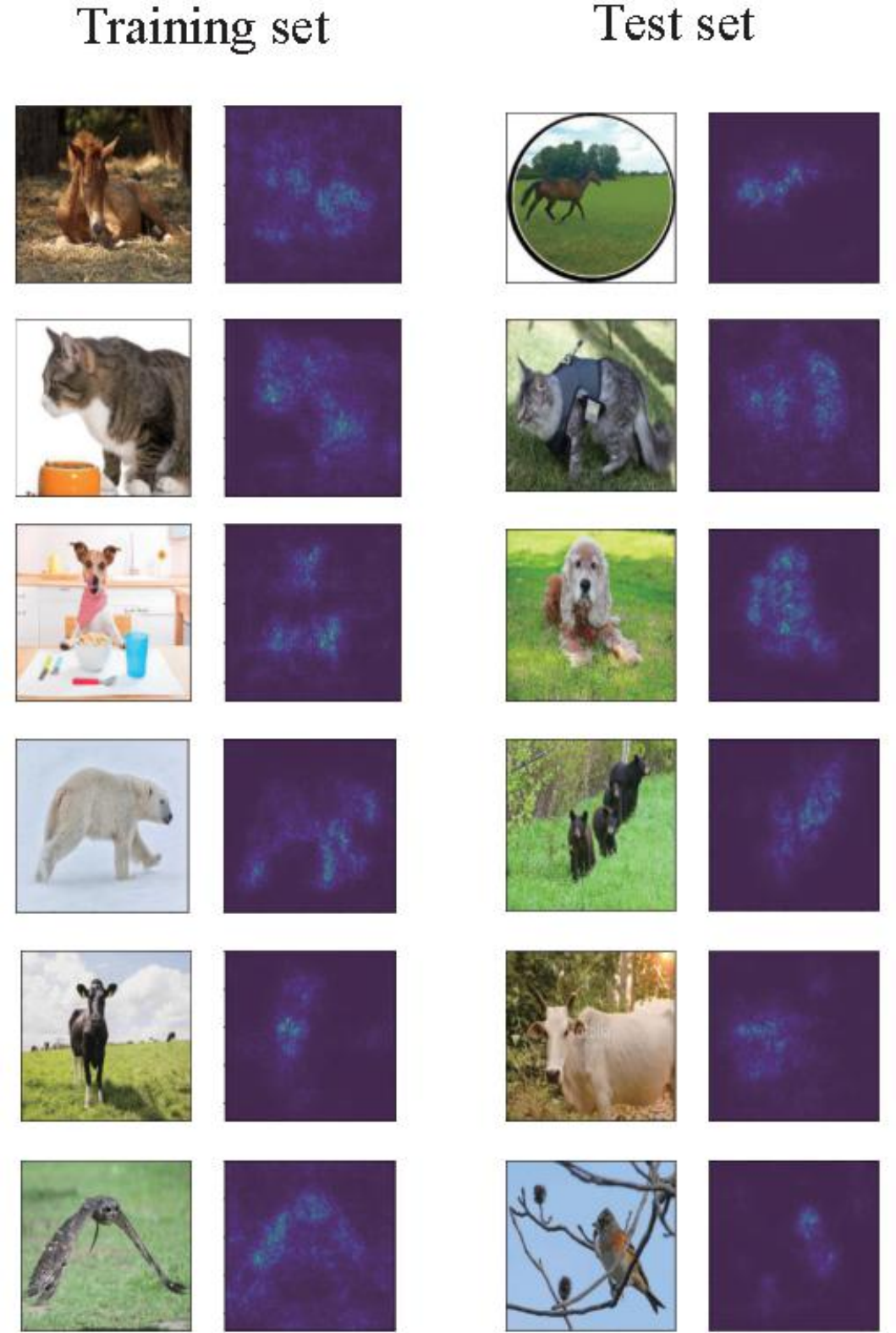}
	\caption{The figure shows the saliency map of NICO-Animal sample.}
\label{heat_map}
\end{figure}

\begin{figure}[ht]\centering
	\includegraphics[width=8cm]{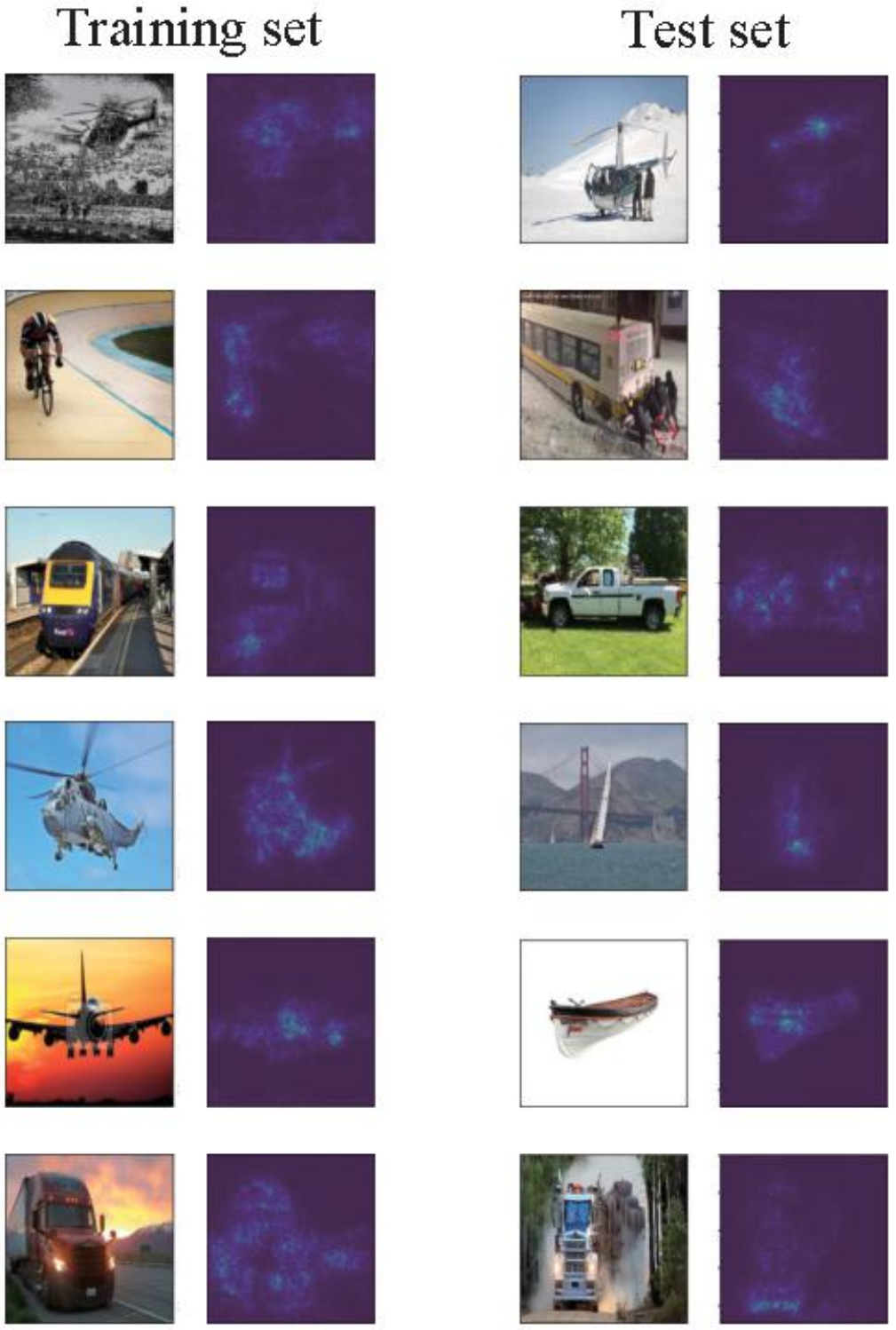}
	\caption{The figure shows the saliency map of NICO-Vehicle sample.}
\label{heat_map_vehicle}
\end{figure}

\begin{table}[ht]
\centering
\caption{D3C and D4C on NICO 8:2}\label{D34C_nico1}
\begin{threeparttable}
\resizebox{0.5\textwidth}{!}{
\begin{tabular}{ccccccc}
\hline\hline
dataset &backbone&\makecell[c]{ Backbone(\%)}&\makecell[c]{Backbone\\+D3C(\%)}&\makecell[c]{Backbone\\+D3C+D4C(\%)}\\
\hline

\multirow{4}*{NICO-Animal}&ResNet18  	  &28.56$\pm$0.30   &{36.79$\pm$0.10} &37.50$\pm$0.05      \\

~&ResNet50	  &39.71$\pm$0.30    &36.43$\pm$0.07    &37.30$\pm$0.10         \\

~&Vgg16       	&48.66$\pm$0.20   &{63.38$\pm$0.04} &\textbf{64.45$\pm$0.08}       \\

~&Vgg19       	&46.11$\pm$0.09    &{61.45$\pm$0.06} &62.85$\pm$0.08      \\ 

\hline
\multirow{4}*{NICO-Vehicle}&ResNet18  	   &42.29$\pm$0.08  &47.82$\pm$0.05 &51.91$\pm$0.06        \\

~&ResNet50	  &55.16$\pm$0.35    &47.82$\pm$0.10 &48.90$\pm$0.10           \\

~&Vgg16       	&58.24$\pm$0.40   &{68.09$\pm$0.14} &69.75$\pm$0.15      \\

~&Vgg19       	&57.60$\pm$0.80  &{70.42$\pm$0.06} &\textbf{71.40$\pm$0.12}      \\ 

\hline\hline
\end{tabular}
}

 \begin{tablenotes}
        \footnotesize
        \item 8 source domains for training and 2 target domains for testing
      \end{tablenotes}
  \end{threeparttable}
  
\end{table}
Previous works have utilized saliency maps to demonstrate the ability of their designed deep learning algorithms to identify foreground objects in NICO data. In this paper, we present saliency maps generated by the VGG16 network enhanced with D3C and D4C method. The saliency maps are shown in Figures \ref{heat_map} and \ref{heat_map_vehicle}.



This paper also compares our method with a series of approaches that do not rely on domain labels. Despite these methods incorporating pre-trained parameters into their backbones, our D3C and D4C methods outperform them when applied to the VGG16 network without the use of pre-trained parameters. The comparison results are recorded in Table \ref{horizontal comparison}.

\begin{table}[ht]
\centering
\caption{Reasoning accuracies on NICO 8:2}\label{horizontal comparison}
\begin{threeparttable}
\begin{tabular}{cccc}
\hline\hline
data set &model&\makecell[c]{average accuracy}&\\
\hline
\multirow{7}*{\makecell[c]{NICO-Animal \\and\\ NICO-Vehicle}}&JiGen    &54.72   \\

~&M-ADA	  &40.78        \\

~&DG-MMLD     	&47.18   \\

~&RSC      	&57.59      \\ 

~&\makecell[c]{ResNet-18(pre-training)} &51.71 \\

~&StableNet   &59.76 \\

~&Vgg16+D3c+D4C   &\textbf{67.10} \\
~&Vgg19+D3c+D4C   &\textbf{67.12} \\
\hline\hline
\end{tabular}

 \begin{tablenotes}
        \footnotesize
        \item 8 source domains for training and 2 target domains for testing
      \end{tablenotes}
  \end{threeparttable}
\end{table}

\subsection{Experiment on  RAVEN}
We tested Lico-Net on the RAVEN and I-RAVEN datasets to showcase its potential. In our exploration of the maximum potential of Lico-Net in solving RPM problems, we decided to discard the use of Sinkhorn distance for probabilistic representations decoding and instead adopted a Transformer-Encoder\cite{Transformer} equipped with learnable class vectors. Initially, we employed the Sinkhorn distance in preliminary experiments to investigate how Lico-Net induces RPM concepts. However, in order to accelerate model convergence and enhance its flexibility, we have chosen the Transformer-Encoder to undertake the task of representation decoding.

Additionally, we conducted experiments combining D3C and D4C with Lico-Net to showcase the effectiveness of the D3C and D4C methods. The D4C method generates 5 negative examples for each RAVEN instance. For a fair comparison, we used the same settings and equipment as the current state-of-the-art (SOTA) method, Triple-CFN\cite{Triple-CFN}, including data volume, optimizer parameters, data augmentation, batch size, and other hyperparameters. 

\begin{table}[h]
\caption{Reasoning accuracies on RAVEN and I-RAVEN.}
\label{RAVEN_IRAVEN_Results}
\centering
\resizebox{\linewidth}{!}{
\begin{tabular}{cccccccccc}
\toprule
\toprule
&\multicolumn{8}{c}{Test Accuracy(\%)}& \\
\cmidrule{2-9}
Model&Average&Center&2 $\times$ 2 Grid&3 $\times$ 3 Grid&L-R&U-D&O-IC&O-IG \\
\midrule
SAVIR-T \cite{SAVIR-T}&94.0/98.1&97.8/99.5&94.7/98.1&83.8/93.8&97.8/99.6&98.2/99.1&97.6/99.5&88.0/97.2\\
\cmidrule{2-9}
SCL \cite{SCL, SAVIR-T}&91.6/95.0&98.1/99.0&91.0/96.2&82.5/89.5&96.8/97.9&96.5/97.1&96.0/97.6&80.1/87.7\\
\cmidrule{2-9}
MRNet \cite{MRNet}&96.6/-&-/-&-/-&-/-&-/-&-/-&-/-&-/-\\
\cmidrule{2-9}
RS-TRAN\cite{RS}&{98.4}/98.7&99.8/{100.0}&{99.7}/{99.3}&{95.4}/96.7&99.2/{100.0}&{99.4}/99.7&{99.9}/99.9&{95.4}/95.4 \\
\cmidrule{2-9}
Triple-CFN\cite{Triple-CFN}&98.9/{99.1}&100.0/{100.0}&99.7/{99.8}&96.2/{97.5}&99.8/{99.9}&99.8/{99.9}&99.9/99.9&97.0/{97.3} \\
\cmidrule{2-9}
Triple-CFN+Re-space\cite{Triple-CFN}&99.4/{99.6}&100.0/{100.0}&99.7/{99.8}&98.0/{99.1}&99.9/{100.0}&99.9/{100.0}&99.9/99.9&98.5/{99.0} \\
\cmidrule{2-9}
Lico-Net&98.7/{99.1}&100.0/{100.0}&99.5/{99.6}&95.9/{97.0}&99.8/{99.9}&99.8/{99.9}&99.9/99.9&96.2/{97.4} \\
\cmidrule{2-9}
Lico-Net+D3C&98.9/{99.3}&100.0/{100.0}&99.7/{99.8}&96.2/{97.5}&99.9/{100.0}&99.9/{100.0}&99.9/99.9&96.8/{98.0} \\
\cmidrule{2-9}
Lico-Net+D4C&99.5/\textbf{99.7}&100.0/\textbf{100.0}&99.7/\textbf{99.8}&98.2/\textbf{99.2}&99.9/\textbf{100.0}&99.9/\textbf{100.0}&99.9/99.9&98.6/\textbf{99.2} \\
\bottomrule
\bottomrule
\end{tabular}
}
\end{table}

An interesting phenomenon observed in Table \ref{RAVEN_IRAVEN_Results} indicates that existing excellent RPM reasoning engines, like Lico-Net, exhibit varying accuracies on the RAVEN\cite{RAVENdataset} and I-RAVEN\cite{I-RAVEN} datasets. I-RAVEN improves the design of negative examples by increasing the number of attribute modifications and the magnitude of deviations, aiming to rectify the unreasonable designs within the option pool of the RAVEN dataset. However, its reasoning logic and problem-solving path have not undergone fundamental changes, and thus its difficulty has not been substantially increased. Surprisingly, I-RAVEN, which does not change in reasoning difficulty, triggers variations in the performance of reasoning engines—a phenomenon that previous works have noticed but failed to explain. However, this paper elucidates the correlation between the quality of negative examples and the precision of learning concepts by reasoning engines, providing a reasonable explanation for this phenomenon, which constitutes one of the theoretical contributions of this paper.

Table \ref{RAVEN_IRAVEN_Results} further demonstrates that Lico-Net, as a reasoning engine that learns probabilistic representations, achieves significantly superior performance compared to conventional reasoning engines based on deterministic representation learning. Furthermore, D4C not only enhances Lico-Net's performance but also, to some extent, bridges the gap in Lico-Net's performance between the RAVEN and I-RAVEN datasets.

\subsection{Experiment on  Bongard-Logo}

We experimented with Lico-Net-Bongard on the Bongard-Logo dataset and explored the enhancement of D3C and D4C algorithms on various Bongard-Logo reasoning engines. D4C generates 5 negative examples. Results are in Table \ref{Bongard_Results}. Notably, D4C excels at generating probabilistic representations of high-scoring negative examples for Triple-CFN and Lico-Net-Bongard, but this can cause gradient explosion during training. Our solution is to use a weighted summation of generated and dataset-provided negative example representations, with a 0.2$:$0.8 ratio. During training, we utilized the Adam optimizer with a learning rate of $10^{-3}$ and a weight decay of $10^{-4}$.

\begin{table}[htbp]
\caption{Reasoning accuracies of models on Bongard-logo.}
\label{Bongard_Results}
\centering
\resizebox{\linewidth}{!}{
\begin{tabular}{ccccccc}
\toprule
&\multicolumn{5}{c}{ Accuracy(\%)}& \\
\cmidrule{2-6}
Model&Train& FF&BA&CM&NV \\
\midrule
MetaOptNet &75.9&	60.3&	71.6&	65.9&	67.5\\
\midrule
ANIL & 69.7 & 56.6 & 59.0 & 59.6 & 61.0\\
\midrule
Meta-Baseline-SC & 75.4 & 66.3 & 73.3 & 63.5 & 63.9 \\
\midrule
Meta-Baseline-MoCo\cite{MoCo} & 81.2 & 65.9 & 72.2 & 63.9& 64.7 \\
\midrule
WReN-Bongard & 78.7 & 50.1 & 50.9 & 53.8 & 54.3 \\
\midrule
SBSD&83.7&75.2&91.5&71.0&74.1\\
\midrule
PMoC&92.0&90.6&97.7&77.3&76.0\\
\midrule
Triple-CFN&93.2&92.0&{98.2}&{78.0}&{78.1}\\
\midrule
\midrule
Lico-Net-Bongard&92.3&91.0&98.3&76.8&77.0\\
\midrule
Lico-Net-Bongard+D3C&{92.9}&{91.3}&98.4&{77.2}&{77.3}\\
\midrule
Lico-Net-Bongard+D4C&{93.2}&92.0&{98.8}&{78.5}&{77.9}\\
\midrule
\midrule
Triple-CFN+D3C&93.9&92.5&98.5&78.2&78.5\\
\midrule
Triple-CFN+D4C&\textbf{{94.3}}&\textbf{92.8}&\textbf{{98.8}}&\textbf{{78.7}}&\textbf{{79.0}}\\
\midrule
\midrule
PMoC+D3C&92.5&91.2&98.0&77.7&76.7\\
\midrule
PMoC+D4C&{93.0}&{92.6}&{98.2}&{{78.0}}&{77.5}\\
\bottomrule
\end{tabular}
}
\end{table}


This paper found that augmentation experiments were conducted on both PMoC \cite{PMoC} and Triple-CFN \cite{Triple-CFN} using techniques like rotation and flipping to expand the Bongard-Logo dataset, without significantly altering concept diversity or proportions \cite{PMoC}. The aim was to explore model performance on a larger dataset. We verified D4C's effectiveness on the augmented dataset through experiments, with results in Table \ref{augmented_Bongard_Results}.

\begin{table}[h]
\caption{Reasoning accuracies of D4C on augmented Bongard-logo.}
\label{augmented_Bongard_Results}
\centering
\resizebox{\linewidth}{!}{
\begin{tabular}{ccccccc}
\toprule
&\multicolumn{5}{c}{ Accuracy(\%)}& \\
\cmidrule{2-6}
Model&Train& FF&BA&CM&NV \\
\midrule
PMoC&94.5&92.6&98.0&78.3&76.5\\
\midrule
Triple-CFN&94.9&93.0&{99.2}&{80.8}&{79.1}\\
\midrule
Lico-Net-Bongard&95.3&94.1&99.3&81.5&78.8\\
\midrule
\midrule
PMoC+D4C&96.0&\textbf{95.2}&99.2&\textbf{84.8}&82.8\\
\midrule
Triple-CFN+D4C&{97.7}&94.8&{99.6}&{82.8}&{83.1}\\
\midrule
Lico-Net-Bongard+D4C&\textbf{97.9}&95.0&\textbf{99.6}&84.1&\textbf{84.0}\\
\bottomrule
\end{tabular}

}
\end{table}

The results in Tables \ref{Bongard_Results} and \ref{augmented_Bongard_Results} indicate that Lico-Net-Bongard exhibits superior performance on the augmented Bongard-Logo problem. This suggests that Lico-Net-Bongard, designed based on probabilistic representation learning, may possess greater network capacity and stronger learning abilities. Especially when combined with the D4C algorithm, Lico-Net-Bongard's performance on the augmented Bongard-Logo problem is particularly outstanding, fully demonstrating the effectiveness and potential of probabilistic representation learning as a methodological guide for deep learning algorithms, and verifying the efficacy of the adversarial learning-based negative example quality improvement algorithm designed in this paper.

\subsection{Experiment on  PGM}
In this paper, we conducted experiments related to Lico-Net on PGM, where D4C still generated 5 negative examples, and recorded the results in Table \ref{PGM_Results}. The data recorded in the table reveals many meaningful phenomena.
\begin{table}[htbp]
\caption{Reasoning accuracies of models on PGM.}
\label{PGM_Results}
\centering
\begin{tabular}{ccc}
\toprule
Model&Test Accuracy(\%) \\
\midrule
SAVIR-T \cite{SAVIR-T}&91.2\\
\midrule
SCL \cite{SCL, SAVIR-T}&88.9\\
\midrule
MRNet \cite{MRNet}&94.5\\
\midrule
RS-CNN\cite{RS}&82.8\\
\midrule
RS-TRAN\cite{RS}&{97.5}\\
\midrule
Triple-CFN\cite{Triple-CFN}&{97.8}\\
\midrule
Triple-CFN+Re-space layer\cite{Triple-CFN}&{98.2}\\
\midrule
\midrule
RS-TRAN+D3C&{97.8}\\
\midrule
RS-TRAN+D4C&{98.3}\\
\midrule
Triple-CFN+Re-space layer+D3C&{98.3}\\
\midrule
Triple-CFN+Re-space layer+D4C&{98.8}\\
\midrule
\midrule
Lico-Net&{97.9}\\
\midrule
Lico-Net+D3C&{98.2}\\
\midrule
Lico-Net+D4C&\textbf{99.0}\\
\bottomrule
\end{tabular}
\end{table}

Firstly, it can be seen from the Table \ref{PGM_Results} that Lico-Net demonstrates superior performance compared to other baseline reasoning engines, which reflects the advantages of learning probabilistic representations of concepts.
Secondly, The combinations of Lico-Net+D3C, Triple-CFN+D3C, and RS-Tran+D3C showed limited improvement in reasoning accuracy, possibly due to the fundamental differences among RPM, NICO, and Bongard-Logo. D3C's approach to improving negative sample quality may be more suitable for addressing concept interference and erroneous coupling issues (preliminary experiments indicate that such issues are particularly prominent in Bongard-Logo and NICO), but RPM problems share a common concept space and attribute recognition mechanism, leading to modest improvements. In contrast, D4C significantly enhanced the performance of these models, supporting our hypothesis that better negative samples can improve the performance of deep learning algorithms. This finding holds significant theoretical and practical importance. Lico-Net has greatly benefited from the support of D4C. We also conducted generalization experiments on Lico-Net, D3C, and D4C, with the results presented in Table \ref{Generalization_PGM_2}.
\begin{table}[h]
\caption{Generalization results of Lico-Net in PGM.}
\label{Generalization_PGM_2}
\centering
\resizebox{\linewidth}{!}{
\begin{tabular}{ccccccccc}
\toprule
&\multicolumn{3}{c}{Accuracy(\%)}  \\
\cmidrule{2-4}
Dataset&Lico-Net &Lico-Net+D3C &Lico-Net+D4C  \\
\midrule
Interpolation & {81.5}&{82.0}&\text{82.9} \\
\midrule
Extrapolation & 18.5 & 18.6 & {18.6} \\
\midrule
Held-out Attribute shape-colour & 12.8 & 13.5 & 13.6 \\
\midrule
Held-out Attribute line-type & 26.0 & 26.5& 27.5\\
\midrule
Held-out Triples & 23.0 & 23.5& 24.3 \\
\midrule
Held-out Pairs of Triples & 44.8 & {45.0}& {45.5} \\
\midrule
Held-out Attribute Pairs& 29.5 & {30.0} & {30.4} \\
\bottomrule
\end{tabular}
}
\end{table}

\subsection{Ablation Study}
The negative example generator in this paper is supported by a step-wise linear attention mechanism, and the performance improvement it brings to the generator needs to be clarified. Therefore, ablation experiments were conducted in this paper. 

We removed the linear attention mechanism from the D4C generator while retaining the process of the transformer-encoder for processing input representations, and conducted relevant experiments again. The results are recorded in the entry labeled ``D4C$^-$" in the Table \ref{ab_PGM_Results}, \ref{ab_augmented_Bongard_Results} and
 \ref{ab_D34C_nico1}.
\begin{table}[htbp]
\caption{Ablation studies on PGM.}
\label{ab_PGM_Results}
\centering
\begin{tabular}{ccc}
\toprule
Model&Test Accuracy(\%) \\
\midrule
RS-TRAN+D4C&{98.3}\\
\midrule
Triple-CFN+Re-space layer+D4C&{98.8}\\
\midrule
Lico-Net+D4C&{99.0}\\
\midrule
\midrule
RS-TRAN+D4C$^-$&{97.9}\\
\midrule
Triple-CFN+Re-space layer+D4C$^-$&{98.5}\\
\midrule
Lico-Net+D4C$^-$&{98.7}\\
\bottomrule
\end{tabular}
\end{table}
\begin{table}[h]
\caption{Ablation studies on augmented Bongard-logo.}
\label{ab_augmented_Bongard_Results}
\centering
\resizebox{\linewidth}{!}{
\begin{tabular}{ccccccc}
\toprule
&\multicolumn{5}{c}{ Accuracy(\%)}& \\
\cmidrule{2-6}
Model&Train& FF&BA&CM&NV \\
\midrule
PMoC+D4C&96.0&{95.2}&99.2&{84.8}&82.8\\
\midrule
Triple-CFN+D4C&{97.7}&94.8&{99.6}&{82.8}&{83.1}\\
\midrule
Lico-Net-Bongard+D4C&97.9&95.0&99.6&84.1&84.0\\
\midrule
\midrule
PMoC+D4C$^-$&95.2&93.9&98.5&82.1&80.0\\
\midrule
Triple-CFN+D4C$^-$&95.5&94.0&{99.3}&{81.0}&{80.2}\\
\midrule
Lico-Net-Bongard+D4C$^-$&96.0&94.2&99.3&83.0&80.8\\
\bottomrule
\end{tabular}
}
\end{table}
\begin{table}[h]
\centering
\caption{Ablation studies on NICO 8:2}\label{ab_D34C_nico1}
\begin{threeparttable}
\resizebox{0.5\textwidth}{!}{
\begin{tabular}{ccccccc}
\hline\hline
dataset &backbone&\makecell[c]{Backbone\\+D3C(\%)}&\makecell[c]{Backbone\\+D3C+D4C(\%)}&\makecell[c]{Backbone\\+D3C+D4C$^-$(\%)}\\
\hline

\multirow{4}*{NICO-Animal}&ResNet18  	  &{36.79$\pm$0.10} &37.50$\pm$0.05   &37.10$\pm$0.04    \\

~&ResNet50	  &36.43$\pm$0.07    &37.30$\pm$0.10      &36.60$\pm$0.08   \\

~&Vgg16       	&{63.38$\pm$0.04} &\textbf{64.45$\pm$0.08}    &{64.00$\pm$0.04}    \\

~&Vgg19       	&{61.45$\pm$0.06} &62.85$\pm$0.08   &{62.10$\pm$0.05}   \\

\hline
\multirow{4}*{NICO-Vehicle}&ResNet18  	   &47.82$\pm$0.05 &51.91$\pm$0.06        &49.32$\pm$0.10\\

~&ResNet50	  &47.82$\pm$0.10 &48.90$\pm$0.10    &48.10$\pm$0.10       \\

~&Vgg16       	&{68.09$\pm$0.14} &69.75$\pm$0.15     &{68.62$\pm$0.18}  \\

~&Vgg19       	&{70.42$\pm$0.06} &\textbf{71.40$\pm$0.12}   &{70.60$\pm$0.20}   \\

\hline\hline
\end{tabular}
}

 \begin{tablenotes}
        \footnotesize
        \item 8 source domains for training and 2 target domains for testing
      \end{tablenotes}
  \end{threeparttable}
  
\end{table}
It can be observed from the experimental results recored in these three tables that the step-wise linear attention mechanism serves as a crucial external memory bank for the D4C negative example generator, providing performance improvements.

\section{Conclusion}

This paper points out that when deep learning is applied to problems that embed complex and deep human concepts, deep learning algorithms should explicitly extract these concepts as probabilistic representations and reason based on these representations. The Lico-Net and Lico-Net-Bongard designed in this paper validate this viewpoint and confirm probabilistic representation learning as a key development direction for deep learning.

More notably, this paper proposes that deep learning models tasked with discriminative tasks are more concerned with the distinctions between positive and negative examples rather than the logic of discriminating between them. Focusing on the distinctions between positive and negative examples is feasible and effective in traditional discriminative tasks. However, when the quality of negative examples is not high enough to strictly define the boundaries of concepts, the performance of the algorithm may be compromised. Therefore, this paper suggests that providing stricter negative examples for deep learning models can effectively improve their performance. 
Consequently, this paper designs the GSCLM algorithm for NICO data and a negative example generator based on a step-wise linear attention mechanism for RPM problems, Bongard-Logo problems, and NICO problems. The experimental results demonstrate the effectiveness of these algorithms, with the D4C generator representing a novel attempt in probabilistic representation learning, hinting at the potential of probabilistic representation learnings in the field of AI.

Thus, this paper advocates that while pursuing the design of algorithms that can extract strong performance from low-quality negative examples, the approach of generating high-quality negative examples and enhancing the contrast between positive and negative examples proposed in this paper is also worthy of attention and consideration.

\newpage

\end{document}